\documentclass[journal]{IEEEtran}
\usepackage{amsmath}

\usepackage{bbding}
\usepackage{pifont}
\usepackage{txfonts}

\usepackage{amsfonts}

\usepackage{algorithmic}
\usepackage{algorithm}
\usepackage{array}
\usepackage{color}
\usepackage{xcolor}
\usepackage[caption=false,font=footnotesize,labelfont=rm,textfont=rm]{subfig}
\usepackage{textcomp}
\usepackage{stfloats}

\usepackage{url}
\usepackage{verbatim}
\usepackage{graphicx}

\usepackage{cite}
\usepackage{epstopdf}

\usepackage{multirow}
\usepackage{booktabs}
\usepackage{newfloat}
\usepackage{listings}
\usepackage{caption}

\usepackage{subfloat}
\usepackage{sidecap}

\usepackage[pagebackref=true,breaklinks=true,letterpaper=true,colorlinks,bookmarks=false]{hyperref}

\begin{document}

\title{Feature Completion Transformer for Occluded Person Re-identification} 

\author{Tao Wang, Mengyuan Liu$^\dagger$, Hong Liu$^\dagger$, Wenhao Li, Miaoju Ban, Tianyu Guo and Yidi Li

    \thanks{$\dagger$ Corresponding author: Mengyuan Liu, Hong Liu.}
    \thanks{T. Wang, M. Liu, H. Liu, W. Li, M. Ban, T Guo and Y Li are with Key Laboratory of Machine Perception, Peking University, Shenzhen Graduate School, China. (E-mail:taowang@stu.pku.edu.cn, mengyuanliu@pku.edu.cn, hongliu@pku.edu.cn, wenhaoli@pku.edu.cn, miaojuban@stu.pku.edu.cn, levigty@stu.pku.edu.cn, yidili@pku.edu.cn).}
    \thanks{This research is supported by the National Natural Science Foundation of China (No. 62073004), Shenzhen Fundamental Research Program (GXWD20201231165807007-20200807164903001, No. JCYJ20200109140410340).}

}

\markboth{IEEE TRANSACTIONS ON MULTIMEDIA}
{Wang \MakeLowercase{\textit{et al.}}: Feature Completion Transformer for Occluded Person Re-identification}
\maketitle

\begin{abstract}
    Occluded person re-identification is a challenging problem due to the destruction of occluders in different camera views. Most existing paradigms focus on visible human body parts through some external models to reduce noise interference. However, the feature misalignment problem caused by discarded occlusions negatively affects the performance of the network. Different from most previous works that discard the occluded regions, we present \underline{F}eature \underline{C}ompletion Trans\underline{former} (FCFormer) that reduces noise interference and complements missing features in occluded parts. Specifically, Occlusion Instance Augmentation is proposed to simulate real and diverse occlusion situations on the holistic image, which enlarges the occlusion samples in the training set and forms aligned occluded-holistic pairs. To reduce the interference of noise, a two-stream architecture is proposed to learn pairwise discriminative features from aligned image pairs, while obtaining self-aligned occluded-holistic feature level sample-label pairs without additional auxiliary models. To complement the features of occluded regions, a Feature Completion Decoder is designed to aggregate possible information from self-generated occluded features in a self-supervised manner. Further, in order to correlate the completion features with identity information, Feature Completion Consistency loss is introduced to enforce the distribution of the generated completion features to be consistent with the real holistic feature distribution. In addition, we propose the Cross Hard Triplet loss to further bridge the gap between completion features and extracting features under the same ID.  Extensive experiments over five challenging datasets demonstrate that the proposed FCFormer achieves superior performance and outperforms the state-of-the-art methods by significant margins on Occluded-Duke dataset.



\end{abstract}


\section{Introduction}
\IEEEPARstart{P}{erson} Re-Identification (Re-ID) involves identifying a person-of-interest across multiple non-overlapping cameras \cite{person_reid}. This task has a wide range of application backgrounds in many fields, such as video surveillance, activity analysis, sport understanding, and tracking.

In the past few years, most of the existing methods mainly focus on the holistic person Re-ID problem\cite{sun2018beyond, shi2022image, shi2020identity}, which assumes that the body of pedestrians is fully visible. However, in real-world scenarios, such as stations, airports, and shopping malls, person images from surveillance cameras can be easily occluded by some obstacles, \textit{e.g.} plants, umbrellas, cars, or other pedestrians, which poses a challenge for holistic Re-ID to identify persons with incomplete and invisible body parts. Therefore, the task of occluded person re-identification \cite{zhuo2018occluded} is proposed to resolve the above problem.

\begin{figure}[t]
    \centering
    \subfloat[Part Occlusion]{
        \centering
        \includegraphics[width=4cm]{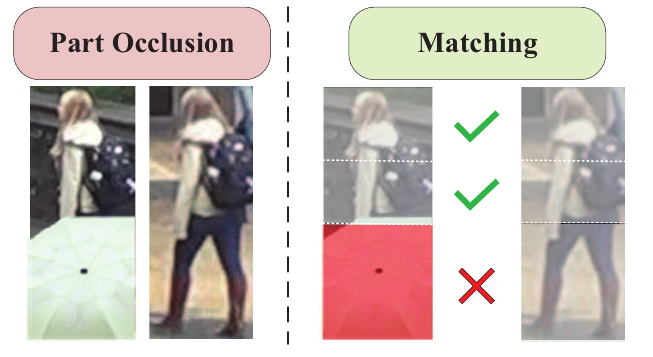}
        \label{part_occ}}
    \hfil
    \subfloat[Complementary Occlusion]{
        \centering
        \includegraphics[width=4cm]{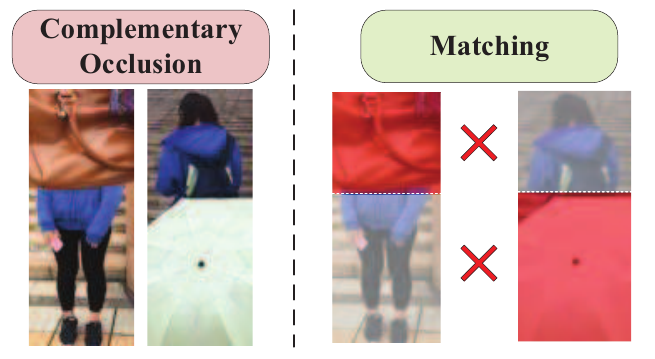}
        \label{com_occ}}

    \subfloat[Feature Completion]{
        \centering
        \includegraphics[width=8.5cm]{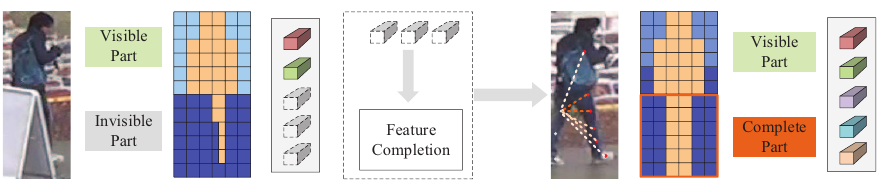}
        \label{fct}}
    \captionsetup{font={footnotesize}}
    \caption{Illustration of part/complementary occlusion and our proposed feature completion paradigm. FCFormer implicitly exploits neighboring region information by using a transformer decoder to recover missing features in occluded regions.}
    \label{occluder}
\end{figure}

Occluded person Re-ID faces more challenges because of the following three problems:
(1) The limited number of occlusion samples in the training dataset \cite{miao2019pose} makes the model sensitive to diverse occlusions. (2) Occlusions will introduce a lot of noise information, which interferes with feature extraction. (3) Occlusions will cause the loss of appearance information, making the extracted features less discriminative and incorrect semantic alignment for matching, as shown in Fig. \ref{part_occ} and Fig. \ref{com_occ}.

To address the above problems, many occluded Re-ID methods are proposed. For the problem (1), some researches \cite{OAMN, DRL_Net} focus on designing occlusion augmentation strategies to improve the robustness of the model. However, these occlusion augmentation strategies still cannot simulate the real environment well. Many studies emphasize the significance of increasing the number of occlusion samples during training, yet fail to address the need for improving the diversity of occlusions. For the problem (2), a large proportion of them exploits additional cues (e.g. pose estimation, semantic parsing or human mask) to indicate non-occluded body parts. For example, PGFA \cite{miao2019pose} directly utilizes pose information to indicate non-occluded body parts on the spatial feature map. PVPM\cite{PVPM} and HoReID \cite{HOReID} use graph-based approaches to model topology information by learning node-to-node or edge-to-edge correspondence to further mine the visible parts. These methods directly adpot external models at the inference stage to extract additional semantic information. However, those approaches are sensitive and error-prone when facing complex backgrounds or severe occlusions. For the problem (3), some methods \cite{GAN_Part, GAN_semantic} attempt to utilize GANs to predict the occluded part at the image level to restore the holistic pedestrian. However, the generated regions are still not convincing, resulting in limited performance. A recent work \cite{RFCnet} performs feature completion on occluded regions by using occluded body part landmarks and utilizing the region based encoder-decoder architecture, which efficiently captures spatial information to recover invisible parts from neighboring features. But this method still needs extra key-points information and predesigned regions in the feature space, which is not flexible enough.

\begin{figure}[t]
    \centering
    \includegraphics[width=0.45\textwidth]{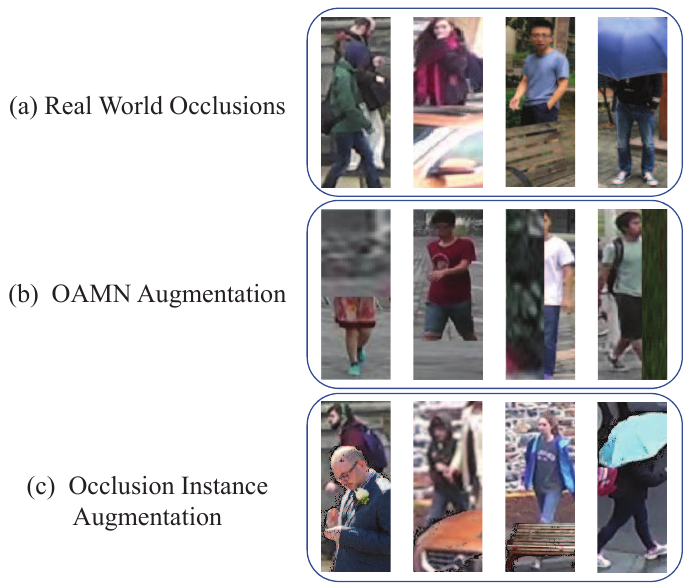}
    \captionsetup{font={footnotesize}}
    \caption{Examples of occluded pedestrians and introducing augmentation. (a) shows the real-world occlusion scenarios. (b) and (c) show augmentation stragy introduced by OAMN\cite{OAMN} and our proposed OIA respectively.}
    \label{fig:Augmentation}
\end{figure}

To simultaneously solve above problems, we propose an \textit{Feature Completion Transformer} (FCFormer) for occluded person Re-ID. Specifically, the proposed method could adaptively complement the features in occluded regions, as shown in Fig. \ref{fct}. Four designs are proposed to alleviate the above three problems. \textbf{Firstly}, to obtain rich occlusion samples, we build an \textit{Occlusion Instances Library} (OIL) that contains 17 classes of occlusion samples obtained from the COCO \cite{COCO} and Occluded-duke \cite{miao2019pose} training sets. Then we propose an \textit{Occlusion Instance Augmentation} (OIA) strategy that produces more diverse occluded training images pairs by pasting image patches from OIL with a specific strategy. As Fig. \ref{fig:Augmentation} shows, compared with the existing occlusion augmentation strategies, such as random erase \cite{Random} and OAMN augmentation \cite{OAMN}, the introduce of rich occlusion samples can better simulate real-world occlusion scenarios. \textbf{Secondly}, a dual-stream encoder paradigm is proposed, which takes the generated holistic-occluded image pairs as input to obtain aligned feature pairs. In general, holistic pedestrian features contain more ID representations than occluded pedestrian features. \textbf{Thirdly}, based on the above prior information, a self-supervised \textit{Feature Completion Decoder} (FCD) module is proposed to complement occluded pedestrian features. FCD has a learnable completion embedding representation, which enables the model to automatically complement missing features under the supervision of holistic pedestrian features without pre-defining human body regions and additional labels.
\textbf{Finally}, to bridge the large feature gap between the occluded scene and the holistic scene, we design a \textit{Cross Hard Triplet} (CHT) loss for metric learning. In addition, we propose a \textit{Feature Completion Consistency} (FC$^2$) loss to explicitly narrow the distribution discrepancy between completion features and holistic features in high-dimensional space, thereby ensuring that FCD could conduct feature completion and train in a self-supervised manner.

The main contributions can be summarized as follows:
\begin{itemize}
    \item [\textbf{(1)}] We develop an Occlusion Instances Library and an Occlusion Instance Augmentation (OIA) strategy for Occluded ReID task, which brings more realistic image-level occlusion enhancement and effectively improve the robustness of the model to various occlusions.
    \item [\textbf{(2)}] A Feature Completion Decoder is proposed to exploit learnable tokens for occluded features completion without the assistance of any external model. Compared to pervious completion based works, our method is more flexible without any predesigned regions.
    \item [\textbf{(3)}] We design a Cross Hard Triplet loss and a Feature Completion Consistency loss to impel model's perception ability and feature completion ability respectively.
    \item [\textbf{(4)}] To validate the effectiveness of our method, we perform experiments on occluded and holistic Re-ID datasets. The results validate the proposed method performs favorably against state-of-the-art methods.
\end{itemize}

\begin{figure*}[t]
    \centering
    \includegraphics[width=1.01\textwidth]{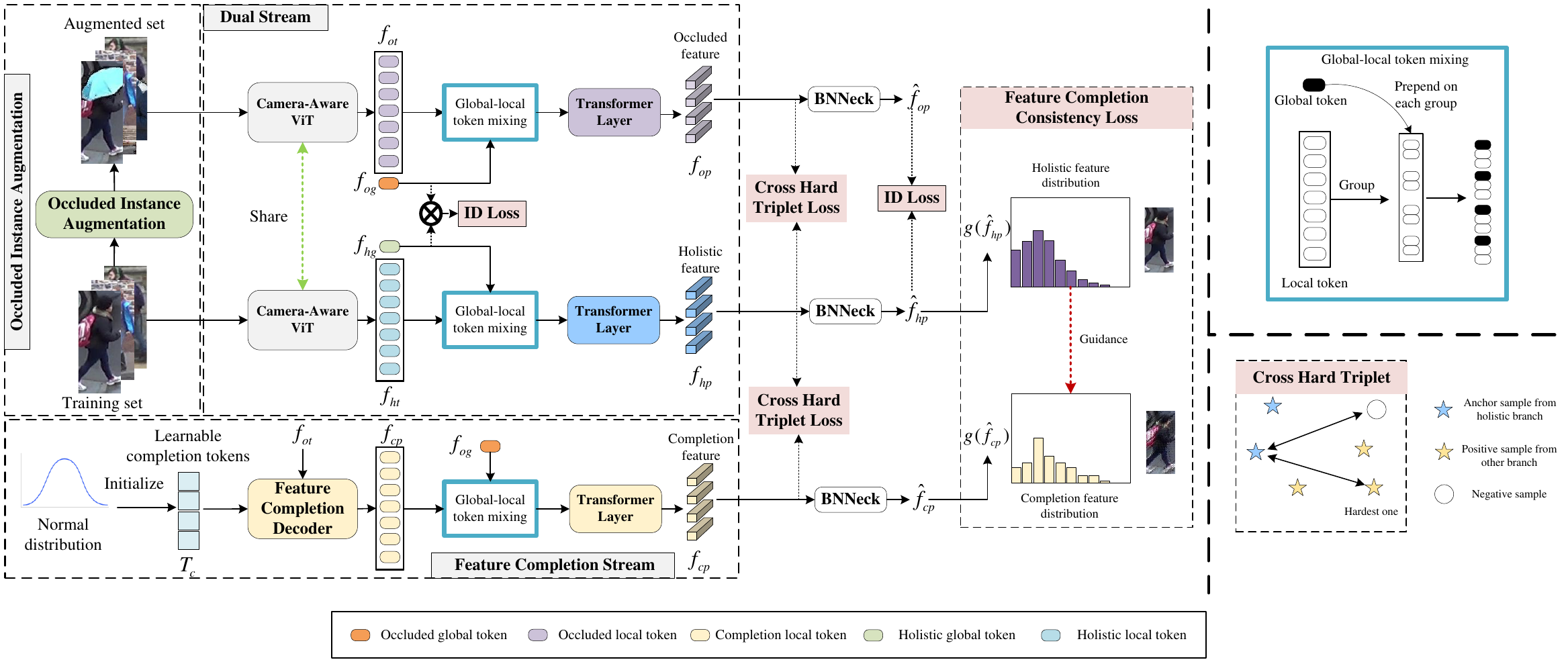}
    \captionsetup{font={footnotesize}}
    \caption{Overall architecture of feature completion transformer (FCFormer). FCFormer consists of three model and two losses, including Occluded Instance Augmentation, Dual stream architecture, Feature Completion Stream, Cross Hard Triplet loss and Feature Completion Consistency loss. The holistic-occluded sample pairs generated from OIA are fed the into dual stream architecture with shared encoder. The non-shared parts of dual stream architecture are used to train the specifical tasks. Then FCD takes the learnable tokens and occluded features as input to recovery holistic features. We propose CHT Loss to allow the model to better perform metric learning among three different modal features (occlusion, holistic, and completion features). At last, FC$^2$ Loss is proposed to guide FCD to generate a completion feature similar enough to the holistic feature. In the test stage, the features from three branches are utilized for retrieval.}
    \label{framework}
\end{figure*}

\section{Related Work}

\subsection{Occluded Person Re-Identification}

Existing methods can be roughly divided into three categories, including part-to-part matching based methods, extra-clues based methods and feature recovery based methods.

Part-to-part matching methods address the occlusion issue by evaluating the similarity between the aligned local features. Sun et al. \cite{sun2018beyond} present a network called Part-based Convolution Baseline (PCB) that uniformly partitions the feature map and learns local features directly. Zhang et al. \cite{alignedreid} aligns features by finding the shortest path among local features. sun et al. \cite{vpm} propose a Visibility-aware
Part Model (VPM), which learns to perceive the visibility of regions by self-supervised learning. Jia et al. \cite{MoS} propose MoS to measure the similarity between person images by using the Jaccard similarity, which formulates the occluded person re-ID as a set matching problem without alignment. The second category is extra-clue based method, which leverages external cues to locate the human body part such as segmentation, pose estimation or body parsing. Song et al.\cite{song2018mask} propose a mask-guided attention model to extract discriminative and robust features invariant to background clutters. Miao et al. \cite{miao2019pose} introduce Pose-Guided Feature Alignment (PGFA) that utilizes gaussian pose heatmap to mine discriminative parts without occlusion. Gao et al. \cite{PVPM} propose a Pose-guided Visible Part Matching (PVPM) model to learn discriminative local features with pose-guided attentions. Wang et al. \cite{HOReID} propose HOReID that utilize GCN to embed the high-order relation and human-topology information between vairous body joints. The last category is feature recovery based methods, which mainly focus on recovering the features of occluded regions. Hou et al. \cite{RFCnet} propose Spatial and Temporal Region Feature Completion (RFC) to recover semantics of occluded regions in feature space for image and video occluded person Re-ID respectively. Yu et al. \cite{Neighbourhood} present that occluded person image features can be reconstructed by its neighborhoods to tackle the problem of occluded person Re-ID. However, all the above recovery-based methods seriously need the support of additional semantic information. Different from the above methods, our method simulates a variety of occlusions, and introduces learnable completion tokens to complete the feature completion of occluded areas in a self-supervised paradigm,

\subsection{Occlusion Augmentation}
The existing person re-identification model is difficult to deal with the occlusion problem, and the interference caused by occlusion limits their robustness. One of the factors is the limited occluded samples in the training set \cite{parallel}, making the model unable to learn the relationship between occlusions and pedestrians.
An effective way to solve this problem is the occlusion augmentation strategy. Currently, occlusion augmentation strategies can be divided into three categories: (1) Random erase. Zhong et al. \cite{Random} propose a method to randomly erase pixel values directly on the image and replace them with random values. This method is simple and helps to reduce the risk of overfitting, but has low generalization. (2) Random cropping and random pasting. Chen et al.\cite{OAMN} randomly cropped a rectangular patch in the training images, and then scaled the cropped area and pasted it in the set four areas randomly. (3) Occluded sample augmentation. Jia et al. \cite{DRL_Net} proposed a method that crops different occlusions from train set and randomly synthesize occlusions for each training batch. Compared with strategy (2), the generated images can better simulate the real occlusion scene, which enable the model implicitly learn more robust features. However, the occlusions in the Occluded-Duke training set are utilized to synthesize, where the categories and number of occlusions are limited, and these methods do not take full advantage of occlusion semantics and paired features brought by augmentation.

\section{Proposed Method}

In this section, we introduce the proposed Feature Completion Transformer (FCFormer) in detail. Firstly, in order to alleviate the model sensitivity problem caused by the small number of occluded samples, we introduce an online data augmentation module named Occlusion Instance Augmentation (OIA) that produces image pairs and occlusion masks (see section \ref{SectionA} for details). Then, a shared dual-stream encoder is proposed to extract pairwise aligned features (see section \ref{SectionB} for details), which could better learn the occlusion relationship from the paired images. A Feature Completion Decoder (FCD) is further proposed to complement human body features in occluded regions (see section \ref{SectionC} for details). The flowchart of our method is shown in Fig. \ref{framework} and the overall FCFormer approach is outlined in Algorithm \ref{alg1}.

\subsection{Occlusion Instance Augmentation}
\label{SectionA}

Most of the existing occlusion augmentation strategies use random cropping to obtain occlusions from randomly sampled images, and stitch the randomly cropped pictures to form an occlusion scene. However, the cropped occlusion has no explicit semantic information and cannot well simulate the occlusion in the real environment. To address above issues, we propose the diversity occlusion instance augmentation strategy. The strategy contains an Occlusion Instance Library (OIL) and an Occlusion Instance Augmentation strategy (OIA).

\begin{figure}[t]
    \centering
    \includegraphics[]{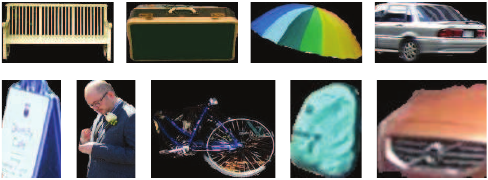}
    \captionsetup{font={footnotesize}}
    \caption{Some occlusion samples from Occlusion Instance Library.}
    \centering
    \label{OIA sample}
\end{figure}

\textbf{Occlusion Instance Library.} In order to better utilize occlusion augmentation to solve the occlusion problem, we propose a general occlusion dataset. By utilizing this dataset, more diverse and realistic occlusion situations can be simulated. We first merge the Occluded-Duke \cite{miao2019pose} training set and the COCO \cite{COCO} training set, then utilize the Mask R-CNN \cite{matterport_maskrcnn_2017} to obtain the instance bounding boxes. In order to avoid the interference of noise information, we erase the pixels irrelevant to the instance. As shown in Fig. \ref{OIA sample}, we manually selected 17 common classes (such as pedestrians, vehicles, bicycles, umbrella, etc.) as occlusion samples with a total of 1000 images.

\begin{table}[htbp]
    \caption{Classes in Occlusion Instance Library.}
    \small
    \centering
    \begin{tabular}{|c|cc|cc|}
        \hline
        \multicolumn{1}{|c|}{\multirow{6}{*}{OIL}} & \multicolumn{2}{c|}{Strong prior set $O_s$} & \multicolumn{2}{c|}{Weak prior set $O_w$}                            \\
        \cline{2-5}
                                                   & car                                         & truck                                     & umbrella & backpack      \\
                                                   & bicycle                                     & motorcycle                                & suitcase & road sign     \\
                                                   & fire hydrant                                & table                                     & kite     & tennis racket \\
                                                   & pedestrian                                  & chair                                     & suitcase & billboard     \\
                                                   & bench                                           & -                                     & -        & -             \\
        \hline
    \end{tabular}
    \label{OIL_P}
\end{table}

\textbf{Occlusion Instance Augmentation.}
Empirically, Some common occlusions have position priors in detected person image (for example, as the Fig. \ref{fig:Augmentation}(a) shows, vehicles are generally in the lower half of the image and are unlikely to appear in other areas of the image). So we determine the augmentation position according to the category of occlusion. As Table \ref{OIL_P} shows, the OIL is divided into two sets: strong position priors set $O_s$ and weak position priors set $O_w$. For the strong position priors set, we align the bottom edge and place them randomly in the horizontal directions. However, for the weak position prior set, its location in the detection box can be relatively random.

\begin{figure}
        \centering
        \includegraphics[width=0.45\textwidth]{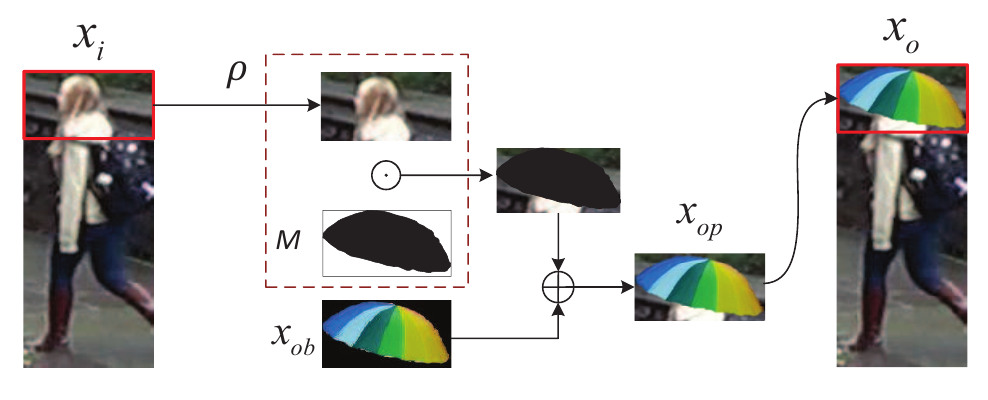}\\
        \captionsetup{font={footnotesize}}
        \caption{Schematic diagram of Occlusion Instance Augmentation.}
        \label{augflow}
    \end{figure}

Specifically, a batch of images can be represented as $X$, each sample can be denoted as $x_i \in X$ and $x_i \in{\mathbb{R}^{H\times{W\times{C}}}} $, where $H, W$ and $C$ represent height, width, and channel dimensions, respectively. Our augmentation scheme has the following few steps: (1) Randomly select an occlusion sample $x_{ob}$ from $OIL$. (2) Randomly Scale the occlusion $x_{ob}$ to 10$\%$$\sim$70$\%$ of image size $H \times W$. It can be described as
    \begin{equation}\label{eq1}
        {\epsilon} = \frac{{\delta}\cdot(H{\times}W)}{h_{o}{\times}w_o},
    \end{equation}
    where $h_o \times w_o$ denotes the size of chosen occlusion sample $x_{ob}$, $\epsilon$ is the ratio of $h_o$ and $w_o$ , $\delta \sim \mathcal{U}(0.1, 0.7)$ is the scaling factor, $\mathcal{U}$ denotes the uniform distribution. (3) Determine the augmentation area. If $x_{ob} \in O_s$, we choose augmentation location $(h, w)$, where $h \in \{H-{\epsilon}h_o, H\}$ and $w \in \{0, W\}$. If $x_{ob} \in O_w$, we randomly put the occlusion sample onto training image $x_i$. Finally, our OIA could be formulated as:
    \begin{equation}\label{eq2}
        x_{op} = \psi(x_{ob}, \epsilon) + M \odot \rho(x_{i}),
    \end{equation}
    where $x_{op}$ denotes the augmented image patches, $\psi(\cdot, \cdot)$ is the resize operation, $\odot$ represents the pixel-wise multiplication, $M$ denotes the occlusion binary mask sampled from resized occlusion $x_{ob}$, and $\rho(\cdot)$ means clipping operation. At last, $x_{op}$ will be paste on the holistic image $x_{i}$. The overall process is shown in Fig. \ref{augflow}.
    Following above precess, we obtain an occluded copies for each training image, and the augmented image is denoted as $x_{o}$.

    \subsection{Dual Stream Architecture}
    \label{SectionB}
    Following methods \cite{Transreid} and \cite{PFD}, we build our feature extractor based on ViT \cite{dosovitskiy2020image}. Given a pair of holistic-occluded images $x_h \in \mathbb{R}^{H\times W\times C}$ and $x_o \in \mathbb{R}^{H\times W\times C}$ from OIA, the shared encoder will split the input image into $N$ non-overlapping patches by using a 2D convolution layer $p(\cdot)$, then the patch embeddings $ E_{p} \in \mathbb{R}^{N \times d}$ can be obtained, where $d$ denotes the embedding dimension. In order to alleviate the impact of camera views, we follow the method in \cite{Transreid} and set a learnable parameter $E_{cm}$ to learn camera viewpoint information. At the same time, a learnable global embedding $E_{g}$ is prepended to the patch embeddings.
    \begin{equation}\label{eq3}
        E_{p} = p(x),
    \end{equation}
    \begin{equation}\label{eq4}
        E_{input} = Concat(E_{g}, E_{p}) + P_{E} + \lambda_{cm} E_{cm},
    \end{equation}
    where $\lambda_{cm}$ is the ratio of camera embeddings, $P_E$ is the learnable position embedding. Then ViT will take $E_{input}$ as input. The output feature for each stream is $f \in {\mathbb{R}^{(N+1)\times C}}$, where $N+1$ denotes the image tokens and one global token, $c$ is the channel dimension. The global tokens $f_{og} \in {\mathbb{R}^{1 \times C}}$ and $f_{hg} \in {\mathbb{R}^{1 \times C}}$ are treated as global feature. Several methods \cite{sun2018beyond, PAT} have proved that fine-grained part features are effective for person re-identification tasks. Thus, the rest of image tokens $f_{ot} \in {\mathbb{R}^{N \times C}}$ and $f_{ht} \in {\mathbb{R}^{N \times C}}$ will be fed into two parallel streams to generate local features.

    \textit{Global-local token mixing:} modeling local relationships from a global perspective can prompt the model to comprehensively understand occlusion relationships. Therefore, drawing inspiration from the local feature generation approach in the TransReID \cite{Transreid}, we perform a global-local token mixing operation, partitioning the image tokens into $M_n$ segments and associating each segment with its corresponding global feature. After obtaining the $M_n$ parts features, we feed them into a non-shared transformer layer and finally get local feature representations $f_{hp} \in {\mathbb{R}^{M_n \times C}}$ and $f_{op} \in {\mathbb{R}^{M_n \times C}}$. Normalized features $\hat{f_{hp}}$ and $\hat{f_{op}}$ can be obtained through a BNNeck\cite{BN}. In short, dual stream architecture consists of a shared ViT backbone and two non-shared transformer layers. The shared backbone can extract general features, while the unshared transformer layer will learn specific patterns for occluded and holistic Re-ID tasks.

    \textit{Loss:} To guarantee that the global and local features are related to the identity and ensure their discriminability, one global and two local classifiers are employed. The model is trained with the following identity loss:
    \begin{equation}\label{eq20}
            \mathcal{L}_{id_g} = -\frac{1}{B}\sum_{i}\log\bigg(\frac{e^{(W_{g}^{y_i})^T\hat{f_{og}^{i}}\hat{f_{hg}^{i}}}}{\sum_{j=1}^{C}e^{(W_{g}^{y_j})^T\hat{f_{og}^{i}}\hat{f_{hg}^{i}}}}\bigg)
    \end{equation}
    \begin{equation}\label{eq21}
        \mathcal{L}_{id_1} = -\frac{1}{B}\sum_{i}\log\bigg(\frac{e^{(W_{h}^{y_i})^T\hat{f_{hp}^{i}}}}{\sum_{j=1}^{C}e^{(W_{h}^{y_j})^T\hat{f_{hp}^{i}}}}\bigg)
    \end{equation}
    \begin{equation}\label{eq22}
        \mathcal{L}_{id_2} = -\frac{1}{B}\sum_{i}\log\bigg(\frac{e^{(W_{o}^{y_i})^T\hat{f_{op}^{i}}}}{\sum_{j=1}^{C}e^{(W_{o}^{y_j})^T\hat{f_{op}^{i}}}}\bigg) 
    \end{equation}
    \begin{equation}\label{eq23}
        \mathcal{L}_{id} = \mathcal{L}_{id_g} + \mathcal{L}_{id_1} + \mathcal{L}_{id_2},
    \end{equation}
    where $B$ denotes the batch size, C is the number of class, and $W_{g}$, $W_{h}$ and $W_{o}$ represent the weight of global, holistic and occluded classifier respectively. The local features $f_{hp}$ and $f_{op}$ have obtained fine-grained features in their respective branches, so we multiply two global features $f_{og}$ and $f_{hg}$ in order to pay more attention to the common parts of paired images at the global feature level.

    \subsection{Feature Completion Stream}

    \label{SectionC}

    \begin{figure}[t]
        \centering
        \includegraphics[width=0.49\textwidth]{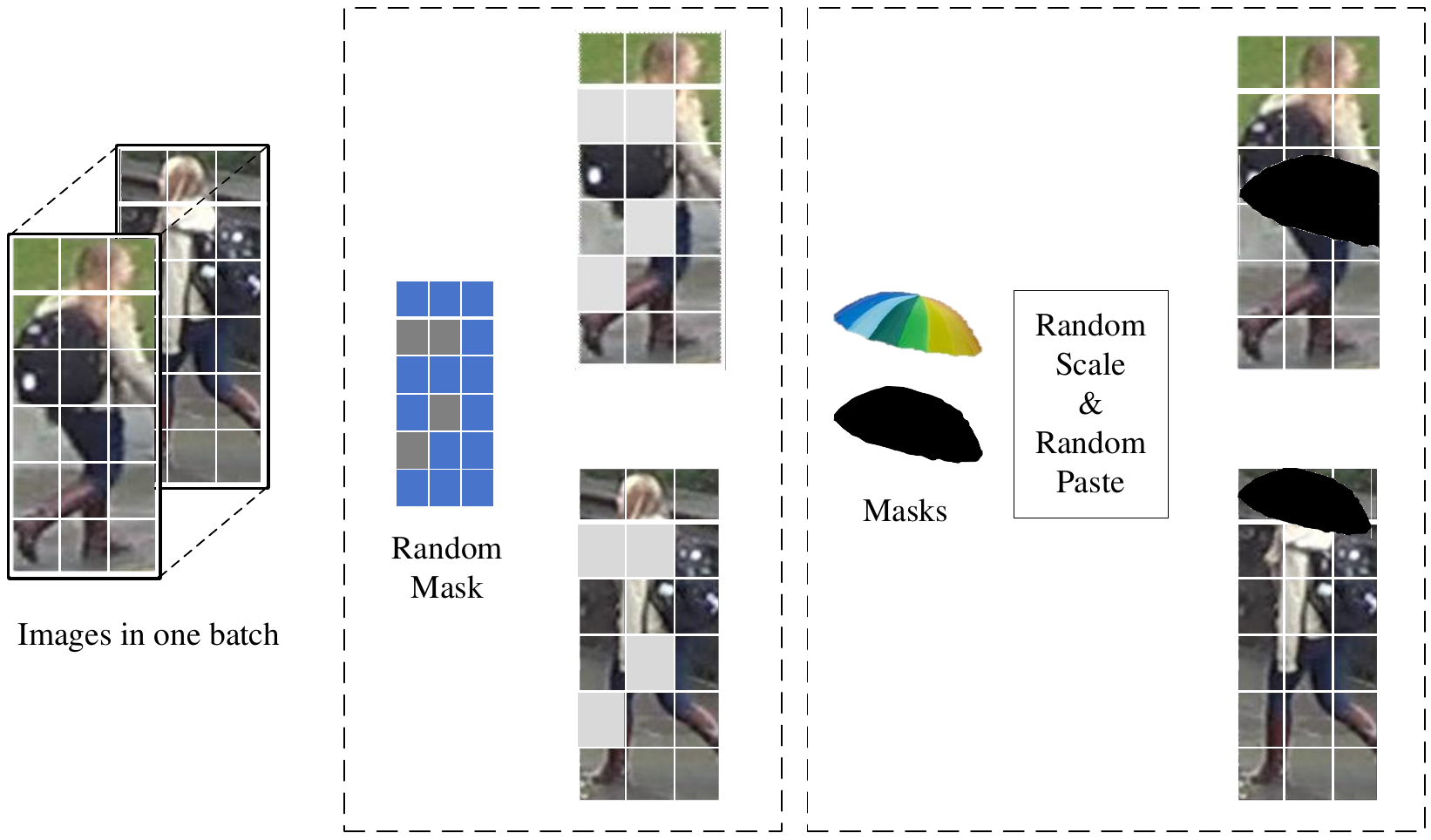}
        \captionsetup{font={footnotesize}}
        \caption{The difference between the feature completion stream and the MAE method in feature recovery. The dotted box on the left is the MAE masking method, which uses a uniform random mask within a batch and lacks diversity. The dotted box on the right is our method, which can generate a variety of occlusion samples within a batch through random scaling and random pasting.}
        \label{fig:tmmaug}
    \end{figure}

    \begin{figure*}[t]
        \centering
        \includegraphics[width=1.0\textwidth]{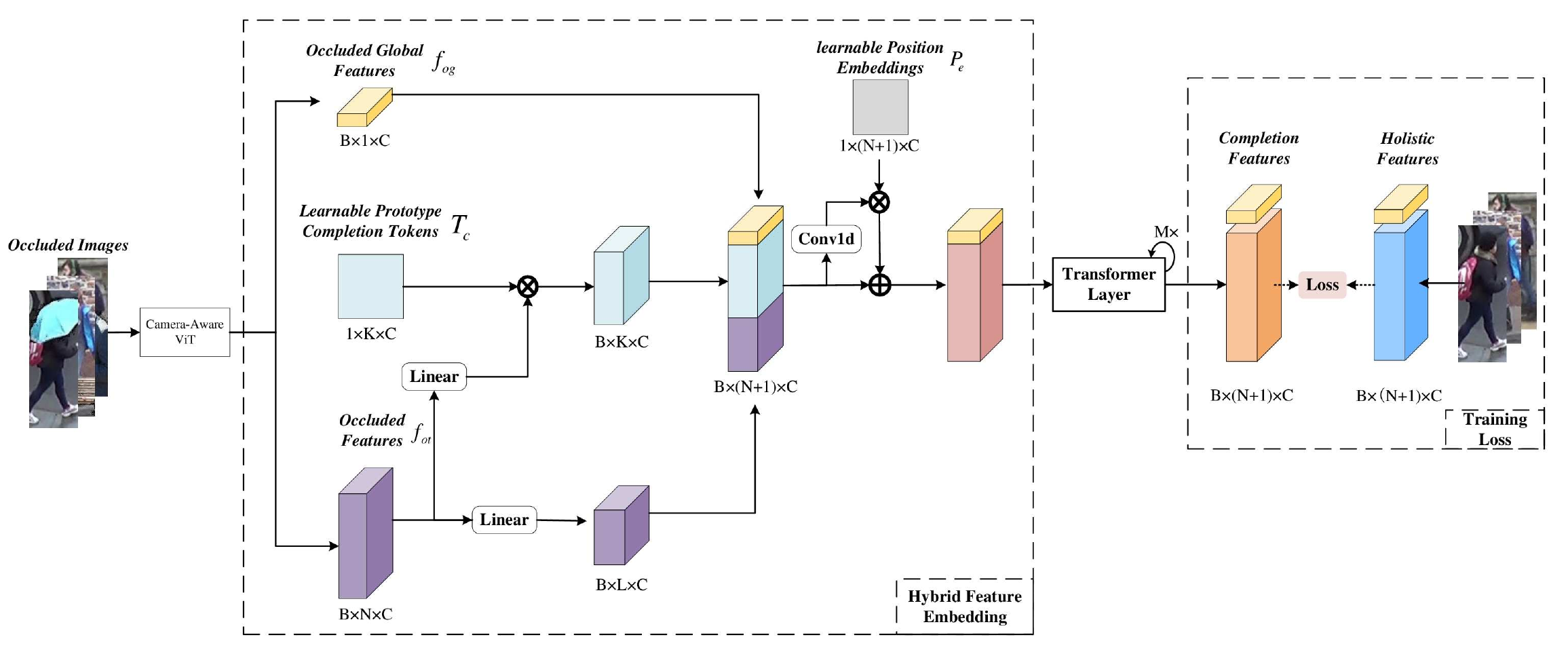}
        \captionsetup{font={footnotesize}}
        \caption{Illustration of the proposed feature completion decoder.}
        \label{fig:FCD}
    \end{figure*}

    Even though the occlusion augmentation strategy and the training method of shared dual-stream network can force the model to better focus on unoccluded parts, the lack of human body part information caused by occlusion has not received much attention. Therefore, we propose a feature completion stream to recover the features of the occluded parts.

    The feautre completion stream consist of a Feature Completion Decoder (FCD) and a transformer layer. The FCD is illustrated in Fig. \ref{fig:FCD}, which consists of a hybrid feature embedding, several transformer layers, and training loss.

    \textit{Hybrid Feature Embedding:} we consider complete pedestrian features to consist of unoccluded visible features and features recovered from occluded regions. Inspired by MAE \cite{he2022masked}, our approach reconstructs the holistic features from the latent occluded features. For each occluded input features, learnable prototype completion tokens $T_{c} \in {\mathbb{R}^{1 \times K \times C}}$ are prepended to aggregate missing feature information from $f_{ot}$. It is worth noting that the position of MAE's mask token is fixed within each batch, which means that each instance can use the same set of mask tokens for feature completion. However, unlike MAE, the occlusion positions of samples generated by OIA are randomly sampled, as shown in Fig.\ref{fig:tmmaug}, therefore we need to conduct completion on each instance. Thus, we fuse the learnable prototype tokens $T_{c} $ with occlusion features to obtain instance-level tokens. Formally,
    \begin{equation}\label{eq5}
        T_b = ({W_{1}}^{T}f_{ot})T_{c},
    \end{equation}
    where $W_{1} \in {\mathbb{R}^{N \times K}}$ is linear projection, $T_{b} \in {\mathbb{R}^{B \times K \times C}}$ is the instance completion tokens. Here, $K = \alpha N$, and $\alpha$ is a hyper-parameter to control the number of tokens. Then we map $f_{ot}$ into $L$ dimension, and concatenate with $T_{b}$. Formally,
    \begin{equation}\label{eq6}
        f_{r} = Concat(f_{og}, T_{b} ,({W_{2}}^{T}f_{ot})),
    \end{equation}
    where $Concat(\cdot)$ is operation of vector combination, $W_{2} \in {\mathbb{R}^{N \times L}}$ is linear projection, and $L = (1-\alpha)N$. $f_{og}$ is prepended to provide global information. Finally, reconstruct feature $f_{r} \in {\mathbb{R}^{B \times (N+1) \times C}}$ is obtained. As mentioned above, due to the randomness of the position of the occluded region, the prepended features need to learn different regions for each instance. However, prepending and concatenation limit position learning and may make the model learn some shortcuts. Therefore, we use a Conv1D to strengthen the association between feature dimensions and use learnable position embeddings to assist the model to encode the token position of each instance. Formally,
    \begin{equation}\label{eq7}
        f_{r} = (W_{3}f_{r})P_{e} + f_{r},
    \end{equation}
    where $W_{3}$ is the parameter of convolution, $P_{e}$ is the positional embedding.

    \textit{Transformer Layers:} the consolidated features are sent to the transformer layers to complement unoccluded body parts. Following the Transformer \cite{vaswani2017attention}, the query, key, and value can be formulated as:
    \begin{equation}\label{eq8}
        Q = W_{q}f_{r}, K = W_{k}f_{r}, V = W_{v}f_{r},
    \end{equation}
    where $W_{q}, W_{k}, W_{v}$ are the weights of linear projections. Through the attention mechanism, the final completion feature can be expressed as:
    \begin{equation}\label{eq9}
        f_m = softmax(\frac{QK^T}{\sqrt{d}})\times V,
    \end{equation}
    \begin{equation}\label{eq10}
        f_{cp} = Wf_m + b.
    \end{equation}

    \textit{Training Loss:} Here we use the holistic feature $f_{ht}$ as the target, which is taken from the complete branch of the dual-stream branch, thus forming a self-supervised training method. The MSE loss function is utilized to drive FCD for training. The training loss can be defined as:
    \begin{equation}\label{eq11}
        \mathcal {L}_{fcd} = \big|\big|f_{cp} - f_{ht}\big|\big|_{2}^{2}.
    \end{equation}

    \subsection{Overall Training Loss}
    We propose a Cross Hard Triplet Loss (CHT) to help the model to better perform metric learning among three different modes of features. (occluded features, holistic features, and complete features).
    Furthermore, In order to successfully complete the occluded pedestrian features, we propose a Feature Completion Consistency Loss (FC$^2$). It is worth noting that this method does not require any additional labels and can be trained in a self-supervised manner.

    \textit{Cross Hard Triplet Loss:} Here, we set holistic feature $f_{hp}$ as anchor, and we want ensure that $f_{hp}$ is closer to all positive samples than other negative samples, which is the same as original triplet loss \cite{schroff2015facenet}. However, most previous models only measure features in a single modality. We find the hardest pair of positive samples and the hardest pair of negative samples among holistic features, occluded features, and complementary features for optimization. Formally,
    \begin{equation}\label{eq12}
        p_{1} = \mathop{\arg\max}\limits_{j} \bigg|f_{hp}^{i} - f_{op}^{j}\bigg|_{+}, p_{2} = \mathop{\arg\max}\limits_{j} \bigg|f_{hp}^{i} - f_{c}^{j}\bigg|_{+},
    \end{equation}
    \begin{equation}\label{eq13}
        n_{1} = \mathop{\arg\min}\limits_{j} \bigg|f_{hp}^{i} - f_{op}^{j}\bigg|_{-}, n_{2} = \mathop{\arg\min}\limits_{j} \bigg|f_{hp}^{i} - f_{c}^{j}\bigg|_{-},
    \end{equation}
    \begin{equation}\label{eq14}
        \begin{split}
            \mathcal {L}_{cht} = \sum_{i}^{B} \max\bigg(\bigg[\big|\big|f_{hp}^{i}-f_{op}^{p_{1}}\big|\big|_{2}^{2}-\big|\big|f_{hp}^{i}-f_{op}^{n_{1}}\big|\big|_{2}^{2} + \alpha \bigg], 0 \bigg) \\
            + \sum_{i}^{B} \max\bigg(\bigg[\big|\big|f_{hp}^{i}-f_{c}^{p_{2}}\big|\big|_{2}^{2}-\big|\big|f_{hp}^{i}-f_{c}^{n_{2}}\big|\big|_{2}^{2} + \alpha\bigg], 0\bigg),
        \end{split}
    \end{equation}
    where $p_1$ and $p_2$ are the index of the hardest positive sample in $f_{op}$ and $f_{c}$ respectively, $n_1$ and $n_2$ are the index of the hardest negative sample in $f_{op}$ and $f_{c}$ respectively. Here, $\alpha$ is a margin hyperparameter.

    \begin{algorithm}[t]
        \caption{Feature Completion Transformer}
        \label{alg1}
        \begin{algorithmic}[1]
            \STATE \textbf{Require} An Occluded training set $\mathcal{X}$, an occlusion Instance library $\mathcal{O}$, a strong prior set $\mathcal{O}_s$, a weak prior set $\mathcal{O}_w$, $\epsilon$ for formula. (\ref{eq1});
            \STATE  $\% Training\enspace stage$
            \FOR{$epoch = 1$ \textbf{to} $num\_epoch$}
            \FOR{mini\_batch $\mathcal{B} \in \mathcal{X}$}
            \FOR{each sample $x \in \mathcal{B}$}
            \STATE Randomly select occlusion sample $x_{ob}$ from $\mathcal{O}$;
            \STATE Scale $x_{ob}$ according to $\epsilon$;
            \IF{$x_{ob} \in \mathcal{O}_s$}
            \STATE Randomly paste $x_{ob}$ onto $x$ at position $(h, w)$, $h \in \{H-{\epsilon}h_o, H\}$ and $w \in \{0, W\}$;
            \ELSE[$x_{ob} \in \mathcal{O}_w$]
            \STATE Randomly paste $x_{ob}$ onto $x$ at position $(h, w)$, $h \in \{0, H\}$ and $w \in \{0, W\}$;
            \ENDIF
            \ENDFOR
            \STATE Extract holistic and occluded features $f_{hp}$ and $f_{op}$ through dual stream architecture;
            \STATE Generate complete feature $f_{c}$ using FCD by taking $T_c$ and $f_{op}$ as input;
            \STATE Compute $\mathcal{L}_{id}$, $\mathcal{L}_{cht}$, $\mathcal{L}_{fc^2}$ by formula. (\ref{eq23}), (\ref{eq14}), (\ref{eq15});
            \STATE Compute final loss in Eq. (\ref{overall loss});
            \STATE Backward to update the Re-ID model;
            \ENDFOR
            \ENDFOR
        \end{algorithmic}
    \end{algorithm}

    \textit{Feature Completion Consistency Loss:} it is almost impossible to complete the ideal holistic feature distribution. So in order to make the distribution of features obtained by completion consistent with the distribution of holistic features, our goal then becomes to minimize the difference between the completed feature distribution and the holistic feature distribution:
    \begin{equation}\label{eq15}
        \mathcal {L}_{fc^2} = \frac{1}{B}\sum_{i}\bigg[p(\hat{f_{c}^{i}})\log p(\hat{f_{c}^{i}})-p(\hat{f_{c}^{i}})\log p(\hat{f_{hp}^{i}})\bigg],
    \end{equation}
    \begin{equation}\label{eq14}
        \begin{split}
            \mathcal {L}_{fc^2} &= H(f_c, f_{hp}) - H(f_c)\\
            &= -\frac{1}{B}\sum_{i}p(\hat{f_{c}^{i}})\log p(\hat{f_{hp}^{i}})-\frac{1}{B}\sum_{i}-p(\hat{f_{c}^{i}})\log p(\hat{f_{c}^{i}})\\
            &= \frac{1}{B}\sum_{i}\bigg[p(\hat{f_{c}^{i}})\log p(\hat{f_{c}^{i}})-p(\hat{f_{c}^{i}})\log p(\hat{f_{hp}^{i}})\bigg]
        \end{split}
    \end{equation}
    where $\hat{f_{c}}$ denotes the feature embedding after BNNeck, $p(\cdot)$ denotes the predicted probability of given features.

    \subsection{Training and Inference}
    The overall process is shown in algorithm \ref{alg1}.
    In the training stage, dual stream architecture and feature completion decoder are trained together with the overall objective loss, which is formulated as:
    \begin{equation}\label{overall loss}
        \mathcal{L} = \mathcal{L}_{id} +\mathcal{L}_{fcd} + \mathcal{L}_{cht} + \mathcal{L}_{fc^2}.
    \end{equation}

    During inference stage, the model is relatively simple, a VIT encoder and a feature completion decoder can perform the inference.

    \section{Experiments}
    \begin{table*}[t]
        \small
        \centering
        \caption{Performance comparison with state-of-the-art methods on Occluded-Duke and P-DukeMTMC. Extra-clue denotes using external model. $\dagger$ means the encoder is with a small step sliding-window setting. Hybird denotes ResNet50 + transformer encoder. $\bigstar$ means the results are reproduced by \cite{FED}.}
        \resizebox{\linewidth}{!}{
            \begin{tabular}{|l|c|c|c|cccc|cc|cc|}
                \hline
                \multicolumn{1}{|l|}{\multirow{2}{*}{Methods}}             & \multicolumn{1}{c|}{\multirow{2}{*}{Publication}} & \multicolumn{1}{c|}{\multirow{2}{*}{Backbone}} & \multicolumn{1}{c|}{\multirow{2}{*}{Extra-clue}} & \multicolumn{4}{c|}{Occluded-Duke} & \multicolumn{2}{c|}{P-DukeMTMC} & \multicolumn{2}{c|}{Occluded-REID}                                                                                     \\    
                \cline{5-12}
                &
                &                                                &                                                  & Rank-1                             & Rank-5                          & Rank-10                            & mAP           & Rank-1        & mAP           & Rank-1          & mAP             \\
                \hline
                Part-Aligned\cite{zhao2017deeply}    &  ICCV $17$        & GoogLeNet                                      & \ding{55}                                        & 28.8                               & 44.6                            & 51.0                               & 44.6          & -             & -             & -               & -               \\    
                PCB\cite{sun2018beyond}             & ECCV $18$               & ResNet50                                       & \ding{55}                                        & 42.6                               & 57.1                            & 62.9                               & 33.7          & 79.4          & 63.9          & 41.3            & 38.9            \\    
                Part Bilinear\cite{PartBilinear}    & ECCV $18$              & GoogLeNet                                      & \checkmark                                       & 36.9                               & -                               & -                                  & -             & -             & -             & -               & -               \\ 
                FD-GAN\cite{fd-gan}     & NIPS $18$                        & ResNet50                                       & \checkmark                                       & 40.8                               & -                               & -                                  & -             & -             & -             & -               & -               \\    
                DSR\cite{DSR}       & CVPR $18$                           & ResNet50                                       & \ding{55}                                        & 40.8                               & 58.2                            & 65.2                               & 30.4          & -             & -             & 72.8            & 62.8            \\ 
                SFR\cite{SFR}        & Arxiv $18$                        & FCN                                            & \ding{55}                                        & 42.3                               & 60.3                            & 67.3                               & 32.0          & -             & -             & -               & -               \\ 
                Ad-Occluded\cite{Ad-occ}    & CVPR $18$                    & ResNet50                                       & \ding{55}                                        & 44.5                               & -                               & -                                  & 32.2          & -             & -             & -               & -               \\ 
                PGFA\cite{miao2019pose}     & ICCV $19$                  & ResNet50                                       & \checkmark                                       & 51.4                               & 68.6                            & 74.9                               & 37.3          & 85.7          & 72.4          & 80.7            & 70.3            \\ 
                PVPM\cite{PVPM}             & CVPR $20$                   & ResNet50                                       & \checkmark                                       & 47.0                               & -                               & -                                  & 37.7          & 85.1          & 69.9          & 70.4            & 61.2            \\ 
                ISP\cite{ISP}   & ECCV $20$                                & ResNet50                                       & \ding{55}                                        & 62.8                               & 78.1                            & 82.9                               & 52.3          & 89.0          & 74.7          & -               & -               \\ 
                HOReID\cite{HOReID} & CVPR $20$                           & ResNet50                                       & \checkmark                                       & 55.1                               & -                               & -                                  & 43.8          & -             & -             & 80.3            & 70.2            \\ 
                SORN\cite{SORN}   & TCSVT $21$                            & ResNet50                                       & \checkmark                                       & 57.6                               & 73.9                            & 79.0                               & 46.3          & -             & -             & -               & -               \\ 
                MoS\cite{MoS}  & AAAI $21$                                & ResNet50                                       & \ding{55}                                        & 61.0                               & 77.4                            & 79.1                               & 49.2          & -             & -             & -               & -               \\ 
                OAMN\cite{OAMN}  & ICCV $21$                              & ResNet50                                       & \ding{55}                                        & 62.6                               & 77.5                            & -                                  & 46.1          & -             & -             & -               & -               \\
                RFCnet\cite{RFCnet}  & TPAMI $21$                         & ResNet50                                       & \checkmark                                       & 63.9                               & 77.6                            & 82.1                               & 54.5          & -             & -             & -               & -               \\
                Pirt\cite{Pirt}   & ACM MM $21$                           & ResNet50-ibn                                   & \checkmark                                       & 60.0                               & -                               & -                                  & 50.9          & -             & -             & -               & -               \\    
                PGFL-KD\cite{PGFL_KD2021}  & ACM MM $21$                   & ResNet50                                       & \checkmark                                       & 63.0                               & -                               & -                                  & 54.1          & -             & -             & 80.7            & 70.3            \\
                QPM\cite{QPM}  & TMM $22$                               & ResNet50                                       & \ding{55}                                        & 64.4                               & 79.3                            & 84.2                               & 49.7          & 89.4          & 74.4          & -               & -               \\
                \hline
                TransReID\cite{Transreid}     & ICCV $21$                 & ViT-B                                          & \ding{55}                                        & 64.2                               & -                               & -                                  & 55.7          & -             & -             & 70.2$^\bigstar$ & 67.3$^\bigstar$ \\ 
                PAT\cite{PAT}      & CVPR $21$                            & Hybrid                                         & \ding{55}                                        & 64.5                               & -                               & -                                  & 53.6          & -             & -             & 81.6            & 72.1            \\ 
                DRL-Net\cite{DRL_Net}  & TMM $22$                             & Hybrid                                         & \ding{55}                                        & 65.8                               & 80.4                            & 85.2                               & 53.9          & -             & -             & -               & -               \\ 
                PFD\cite{PFD}   & AAAI $22$                                & ViT-B                                          & \checkmark                                       & 67.7                               & 80.1                            & 85.0                               & 60.1          & -             & -             & 79.8            & 81.5            \\ 
                FED\cite{FED}   & CVPR $22$                                & ViT-B                                          & \ding{55}                                        & 68.1                               & -                               & -                                  & 56.4          & -             & -             & \textbf{86.3}   & 79.3            \\ 
                FRT\cite{xu2022learning}   & TIP $22$                                &  Hybrid                                          & \checkmark                                      & 70.7                            & -                               & -                                  & \textbf{61.3}          & -             & -             & 81.6   & 72.1            \\ 
                \textbf{FCFormer \textit{(Ours)}}     & -                     & ViT-B                                          & \ding{55}                                        & \textbf{71.3}                      & \textbf{84.1}                   & \textbf{87.1}                      & 60.9 & \textbf{91.5} & \textbf{80.7} & 84.9            & \textbf{86.2}   \\ 
                \hline
                TransReID$^\dagger$\cite{Transreid}   & ICCV $21$          & ViT-B                                          & \ding{55}                                        & 66.4                               & -                               & -                                  & 59.2          & -             & -             & -               & -               \\ 
                PFD$^\dagger$\cite{PFD}  & AAAI $22$                      & ViT-B                                          & \checkmark                                       & 69.5                               & -                               & -                                  & 61.8          & -             & -             & 81.5            & 83.0            \\ 
                \textbf{FCFormer$^\dagger$ \textit{(Ours)}}  & -              & ViT-B                                          & \ding{55}                                        & \textbf{73.0}                      & \textbf{84.9}                   & \textbf{88.6}                      & \textbf{63.1} & \textbf{92.4} & \textbf{82.5} & 83.6            & \textbf{85.7}   \\ 

                \hline
                PGFA\cite{miao2019pose}  + Re-ranking   & ICCV $19$                    & ResNet50                                       & \checkmark                                       & 52.4                               & 68.6                            & 74.9                               & 46.8          & -             & -             & -               & -               \\ 
                HOReID\cite{HOReID}  + Re-ranking      & CVPR $20$                     & ResNet50                                       & \checkmark                                       & 58.3                               & -                               & -                                  & 49.2          & -             & -             & -               & -               \\ 
                Pirt\cite{Pirt}  + Re-ranking   & ACM MM $21$                            & ResNet50-ibn                                   & \checkmark                                       & 62.1                               & -                               & -                                  & 59.3          & -             & -             & -               & -               \\ 
                PFD\cite{PFD} + Re-ranking      & AAAI $22$                           & ViT-B                                          & \checkmark                                       & 71.7                               & 79.4                            & 82.3                               & 71.7          & -             & -             & - -             & -               \\ 
                \textbf{FCFormer \textit{(Ours)}} {+ Re-ranking}   & -            & ViT-B                                          & \ding{55}                                        & \textbf{76.8}                      & \textbf{85.7}                   & \textbf{88.4}                      & \textbf{75.8} & 89.1          & \textbf{85.3} & 84.6            & \textbf{86.9}   \\ 
                \textbf{FCFormer$^\dagger$ \textit{(Ours)}} {+ Re-ranking}   & -  & ViT-B                                          & \ding{55}                                        & \textbf{79.4}                      & \textbf{86.7}                   & \textbf{89.3}                      & \textbf{77.2} & 90.6          & \textbf{87.1} & 84.4               & \textbf{86.7}      \\
                \hline
            \end{tabular}}
        \label{Occluded_Results}
    \end{table*}

    \subsection{Datasets and Evaluation Metrics}
    To illustrate the effectiveness of our method, we evaluate our method on five Re-ID datasets for two tasks including occluded person Re-ID, and holistic person Re-ID.

    \textbf{Occluded-Duke} \cite{miao2019pose} consists of 15,618 training images of 702 persons, 2,210 occluded query images of 519 identities and 17,661 gallery images of 1110 persons. It is a subset of DukeMTMC-reID \cite{zheng2017unlabeled}, which is currently the most challenging dataset due to its complex scene.

    \textbf{P-DukeMTMC} \cite{zhuo2018occluded} is the subset of the DukeMTMC-reID, which consists of 12,927 training images with 665 identities, 2163 occluded images for query and 9053 images without occlusion for gallery.

    \textbf{Occluded-REID} \cite{zhuo2018occluded} is captured by the mobile phone, which consist of 2,000 images of 200 occluded persons. Each identity has five full-body person images and five occluded person images with different types of severe occlusions.

    \textbf{Market-1501} \cite{zheng2015scalable} contains 1,501 identities observed from 6 camera viewpoints, 12,936 training images of 751 identities, 19,732 gallery images and 2,228 queries of 750 persons.

    \textbf{DukeMTMC-reID} \cite{zheng2017unlabeled} contains 36,411 images of 1,404 identities captured from 8 camera viewpoints. It contains 16,522 training images, 17,661 gallery images and 2,228 queries.

    \textbf{Evaluation Metrics.} We adopt Cumulative Matching Characteristic (CMC) curves and mean average precision (mAP) to evaluate different Re-ID models.

    \subsection{Implementation Details}
    Our encoder is followed TransReID \cite{Transreid}, which consists of 12 transformer layers. The initial weights of encoder are pre-trained on ImageNet-21K and then finetuned on ImageNet-1K. During the training and testing time, input images are both resized to 256 $\times$ 128. The training images are augmented with random horizontal flipping, and padding. The number of the number of part token $M_N$ is set to 4 for Occluded-Duke dataset (Details are described in Sec. \ref{ablation_section} ). The number of decoder layer is set to 2. And the number of token $\alpha$ in FCD is 0.7. The hidden dimension $D$ is set to 768. The transformer decoder is same with \cite{vaswani2017attention}. The batch size is set to 64 with 4 images per ID. The learing rate is initialized at 0.008 with cosine learning rate decay. We conduct all experiments on one RTX 3090 GPU.

    \subsection{Comparison with the State-of-the-Art Models}
    We compare FCFormer with the state-of-the-art methods on seven benchmarks including occluded person ReID and holistic person ReID.

    \textbf{Results on Occluded datasets.}
    Table \ref{Occluded_Results} shows the results on three occluded datasets, \textit{i.e.}, Occluded-Duke and P-DukeMTMC. As the table shows, two classes of methods will be compared: CNN based ReID methods and Transformer based ReID methods. The results show that the proposed FCFormer can consistently achieve competitive performance on occluded datasets. It is worth noting that some previous methods adopt additional models, such as PFD \cite{PFD} and HOReID \cite{HOReID}, both utilize skeleton topology information to cluster unoccluded features, thus achieving competitive results. In comparison, FCFormer achieves the best result with Rank-1 accuracy of 71.3\% and mAP of 60.9\% on the challenging Occluded-Duke dataset without any external clues, which outperforms the previous state-of-the-art methods by a large margin (at least +3.2\%Rank-1/+0.8\%mAP). On the P-DukeMC dataset, our FCFormer achieves 91.5\% Rank-1 accuracy and 80.7\% mAP, surpassing the state-of-the-art model QPM \cite{QPM} by 2.1\% and 6.3\% in terms of Rank-1 accuracy and mAP respectively. In addition, FCFormer is flexible and scalable. We change the step of sliding-window, and also introduce the Re-ranking technology. As the table shows, FCFormer$^\dagger$ achieves 73.0\%/63.1\% on Rank-1/mAP on Occluded-Duke dataset by simply reducing the step size, surpassing others by at least 4.9\%/3\% in terms of Rank-1 and mAP respectively. Further, with the help of re-ranking, our model FCFormer$^\dagger$ + Re-ranking achieves the highest results by far, reaching 79.4\% Rank-1 and 77.2\% mAP.

    \textbf{Comparison with other recover-based models.}
    RFCnet\cite{RFCnet} and FRT\cite{xu2022learning} are methods based on feature recovery. Compared with these two methods, our thought is somewhat different from them. RFCnet uses a pose estimator to divide pedestrian images into blocks, encodes these regional features, and then clusters them. Finally, the clustered cluster features are decoded and assigned to each human body region to achieve feature recovery. The idea of FRT is similar to RFCnet. It can be said that FRT is a continuation of RFCnet. RFCnet uses nearby region features for feature recovery in a single image, while FRT uses the features of k-nearest instances for recovery. Compared with these two methods, our method induces a new paradigm of learning feature recovery directly from the constructed occluded-holistic pairs without the need for additional pose estimation models as well as feature ranking, and achieves excellent results.

    \begin{table}[t]
        \small
        \centering
        \caption{Performance comparison with state-of-the-art models on Market-1501 and DukeMTMC-reID datasets.}
        \resizebox{\linewidth}{!}{
            \begin{tabular}{l|c|cc|cc}
                \hline
                \multicolumn{1}{l|}{\multirow{2}{*}{Methods}} & \multicolumn{1}{c|}{\multirow{2}{*}{Publication}} & \multicolumn{2}{c|}{Market-1501} & \multicolumn{2}{c}{DukeMTMC}                                 \\
                \cline{3-6}
                                                              & & Rank-1                           & mAP                          & Rank-1        & mAP           \\
                \hline
                PCB\cite{sun2018beyond}  & ECCV $18$         & 92.3                             & 77.4                         & 81.8          & 66.1          \\
                DSR\cite{DSR}  & CVPR $18$                   & 83.6                             & 64.3                         & -             & -             \\
                SPReID\cite{spreid}  &CVPR $18$             & 92.5                             & 81.3                         & 84.4          & 70.1          \\
                BOT\cite{BOT} & CVPRW $19$                  & 94.1                             & 85.7                         & 86.4          & 76.4          \\
                MVPM\cite{MVPM}   &ICCV $19$                 & 91.4                             & 80.5                         & 83.4          & 70.0          \\
                SFT\cite{sft}  &ICCV $19$                   & 93.4                             & 82.7                         & 86.9          & 73.2          \\
                CAMA\cite{CAMA}  &CVPR $19$                 & 94.7                             & 84.5                         & 85.8          & 72.9          \\
                IANet\cite{IAnet}  &CVPR $19$                 & 94.4                             & 83.1                         & 87.1          & 73.4          \\
                P$^2$Net\cite{p2net} & ICCV $19$             & 95.2                             & 85.6                         & 86.5          & 73.1          \\
                PGFA\cite{miao2019pose} &CVPR $19$            & 91.2                             & 76.8                         & 82.6          & 65.5          \\
                AANet\cite{AAnet}  &CVPR $19$                 & 93.9                             & 82.5                         & 86.4          & 72.6          \\
                Circle\cite{circle} &CVPR $20$                & 94.2                             & 84.9                         & -             & -             \\
                HOReID\cite{HOReID}  &CVPR $20$               & 94.2                             & 84.9                         & 86.9          & 75.6          \\
                Pirt\cite{Pirt}    & ACM MM $22$                & 94.1                             & 86.3                         & 88.9          & 77.6          \\
                \hline
                TransReID\cite{Transreid} &ICCV $21$         & 95.0                             & 88.2                         & 89.6          & 80.6          \\
                PAT\cite{PAT} &CVPR $21$                     & 95.4                             & 88.0                         & 88.8          & 78.2          \\
                DRL-Net\cite{DRL_Net} & TMM $22$             & 94.7                             & 86.9                         & 88.1          & 76.6          \\
                PFD\cite{PFD}   & AAAI $22$                  & \textbf{95.5}                    & \textbf{89.6}                & \textbf{90.6} & \textbf{82.2} \\
                FED\cite{FED}   & CVPR $22$                 & 95.0                             & 86.3                         & 89.4          & 78.0          \\
                FCFormer ($Ours$)   & -                           & 95.0                             & 86.8                         & 89.7          & 78.8          \\
                \hline
            \end{tabular}}
        \label{holistic result}
    \end{table}

    \textbf{Results on Holistic ReID datasets.}
    We conduct experiments on two holistic ReID datasets including Market-1501, and DukeMTMC-reID. Table \ref{holistic result} shows the results on Market-1501 and DukeMTMC-reID datasets. Specifically, our method FCFormer achieves comparable performance (95.0\%/86.8\% Rank-1 accuracy and 89.7\%/78.8\% mAP, respectively) on Market-1501 and DukeMTMC-reID datasets. As shown in the results, PFD achieves better results than ours on the holistic datasets. This is mainly because PFD incorporates skeleton information during its encoding and decoding process. Moreover, PFD uses more parts for feature retrieval, and the feature dimension is larger, so the features are more fine-grained. In addition, FCFormer applies a shared backbone, the occlusion branch in the Dual Stream Architecture will affect the backpropagation of the complete feature. Our FCFormer, although not designed for holistic person re-identification, still outperforms CNN-based methods without the help of external models and has a small gap with current state-of-the-art models, illustrating the robustness of FCFormer.

    \subsection{Performance under Transfer setting}
    In order to further vertify the effectiveness of the method, we follow the methods \cite{HOReID, PVPM, QPM} that adopts Market-1501 or MSMT17 as training set, then directly evaluate on occluded dataset (Occluded-REID and P-DukeMTMC). A number of methods \cite{HACNN, OSNet, IDE, Part_bili, sun2018beyond, ISP, miao2019pose, PVPM, QPM} are involved in the comparison. Table \ref{Occluded_Results} and \ref{transfer_setting} show that FCFormer achieves the competitive Rank-1 accuracy and mAP on Occluded-REID dataset and P-DukeMTMC dataset respectively. On the P-DukeMTMC dataset, our FCFormer outperforms QPM by 1.0\% and 8.3\% in terms of Rank-1 accuracy and mAP respectively. On the Occluded-REID dataset, FCFormer produces the comparable results with FED \cite{FED}, achieving 84.9\% Rank-1. We fail to achieve the highest Rank-1 performance on Occluded-REID for the following reasons: First, the Transformer has poor cross-domain generalization ability on small datasets.
    Secondly, we use the same backbone as TransReID, which contains some dataset-specific tokens, further limiting the generalization ability. However, it is worth noting that our method achieves 86.2\% mAP on Occluded-REID and 39.4\% mAP on P-DukeMTMC, surpassing previous model by at least 4.7\% and 8.3\% on Occluded-REID and P-DukeMTMC respectively. The reason for this is that the completion feature participates in the feature distance calculation, which helps the average accuracy.

    \subsection{Recover from Complement Features}
\begin{figure*}[t]
        \centering
        \includegraphics[width=0.75\textwidth]{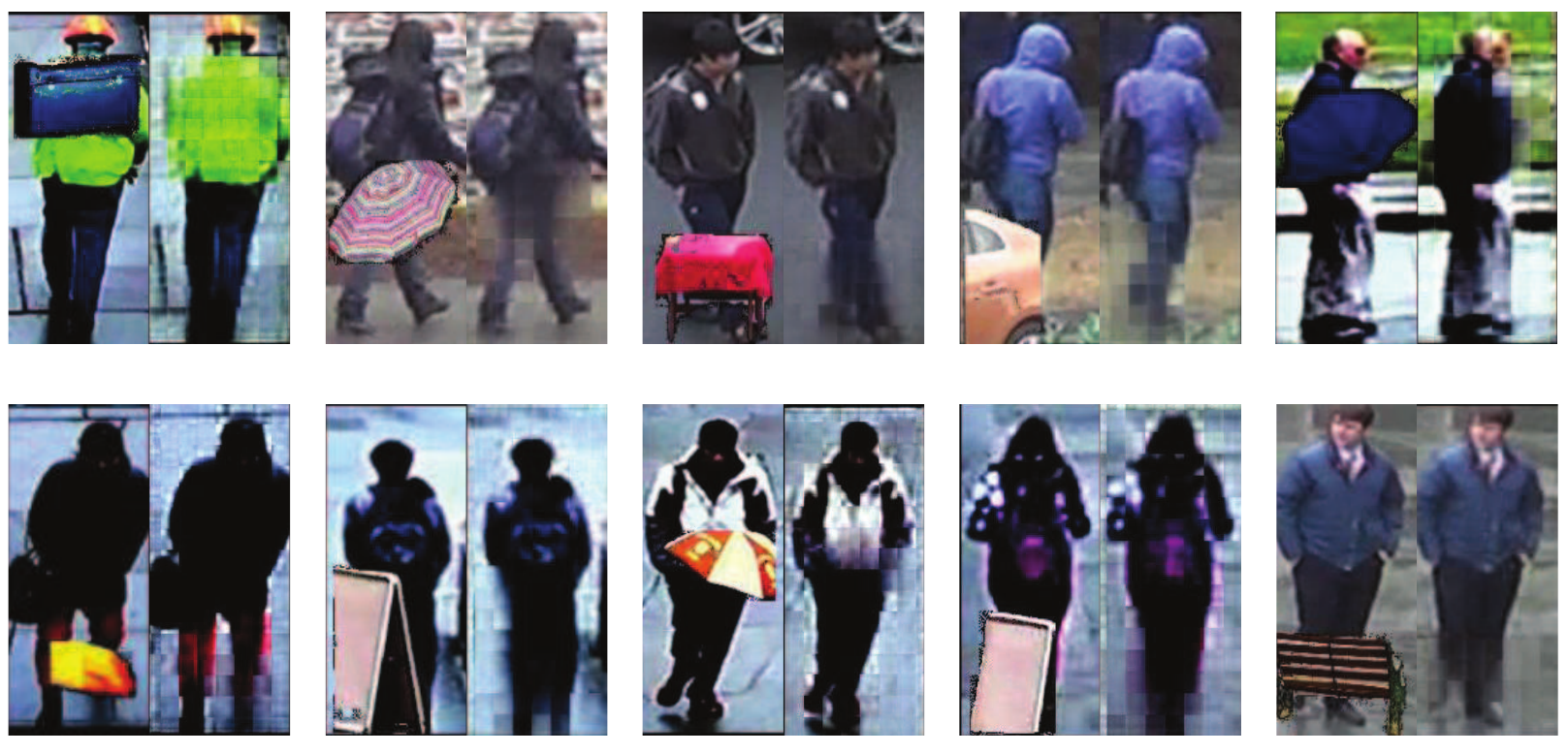}
        \captionsetup{font={footnotesize}}
        \caption{Complementing results. For each pair, we show the augmentated image (left) and reconstruction from FCD feature (right). }
        \label{recover_sample}
    \end{figure*}

    We design an image decoder to recover holistic pedestrian images from latent completion features, demonstrating that learnable completion tokens could learn missing features. We fix all the weights of the propsed network and only train the image decoder to perform the image reconstruction task. The loss function computes the mean squared error (MSE) between the reconstructed and original images in the pixel level. Inconsistent with the decoder in MAE\cite{he2022masked}, The model infers occluded patches instead of masked patches to produce reasonable output by predicting the pixel value. Qualitative results can be found in Fig. \ref{recover_sample}. We can see that the learnable completion tokens use saliency information of identified neighbor regions to compensate for missing features, which demonstrate the effectiveness of feature completion decoder.

    \begin{table}[t]
        \small
        \centering
        \caption{Performance comparisons on P-DukeMTMC under transfer setting.}
        \label{P-duke_transfer}
        \resizebox{\linewidth}{!}{\begin{tabular}{l|c|cccc}
                \hline
                Methods (Market $\to$ P-Duke)       & Publication                               & Rank-1        & Rank-5        & Rank-10       & mAP           \\
                \hline
                IDE\cite{IDE}   &  CVPR $17$                 & 36.0          & 49.3          & 55.2          & 19.7          \\
                HACNN\cite{HACNN} &  CVPR $18$                & 30.4          & 42.1          & 49.0          & 17.0          \\
                PCB\cite{sun2018beyond} &  ECCV $18$            & 43.6          & 57.1          & 63.3          & 24.7          \\
                OSNet\cite{OSNet}  & ICCV $19$              & 33.7          & 46.5          & 54.0          & 20.1          \\
                Part Bilinear\cite{PartBilinear}  & ECCV $18$ & 39.2          & 50.6          & 56.4          & 25.4          \\
                ISP*\cite{ISP}   &ECCV $20$                & 46.3          & 56.9          & 60.8          & 26.4          \\
                PGFA$^+$\cite{miao2019pose}  &TNNLS $21$    & 48.2          & 59.6          & 65.8          & 26.8          \\
                PVPM\cite{PVPM} &CVPR $20$                 & 51.5          & 64.4          & 69.6          & 29.2          \\
                QPM\cite{QPM} &TMM $22$                    & 57.3          & \textbf{69.9} & \textbf{75.5} & 31.1          \\
                \hline
                \textbf{FCFormer} (Market $\to$ P-Duke) &-    & \textbf{58.3} & 69.7          & 74.4          & \textbf{39.4} \\
                \textbf{FCFormer} (MSMT17 $\to$ P-Duke) &-     & \textbf{58.2} & 69.8          & 74.6          & \textbf{41.4} \\
                \hline
            \end{tabular}}
        \label{transfer_setting}

    \end{table}

    \begin{table*}[t]
        \small
        \centering
        \caption{Ablation study over Occluded-Duke.}
        \begin{tabular}{c|cccccc|cccc}
            \hline
            Index & Baseline   & Dual-stream & OIA        & FCD        & CHT loss   & FC$^2$ loss & R-1                          & R-5  & R-10 & mAP                          \\
            \hline
            \hline
            1     & \checkmark &                &            &            &            &             & 59.3                         & 76.5 & 82.2 & 50.0                         \\
            2     & \checkmark &                & \checkmark &            &            &             & 63.4(\textcolor{red}{+4.1})  & 77.1 & 82.6 & 53.2(\textcolor{red}{+3.2})  \\
            3     & \checkmark & \checkmark     & \checkmark &            &            &             & 66.5(\textcolor{red}{+7.2})  & 78.6 & 83.6 & 55.7(\textcolor{red}{+5.7})  \\
            4     & \checkmark &                & \checkmark & \checkmark &            &             & 65.1(\textcolor{red}{+5.8})  & 77.9 & 83.1 & 54.8(\textcolor{red}{+4.8})  \\
            5     & \checkmark & \checkmark     & \checkmark & \checkmark &            &             & 67.3(\textcolor{red}{+8.0})  & 79.5 & 84.1 & 56.9(\textcolor{red}{+6.9})  \\
            6     & \checkmark & \checkmark     & \checkmark &            & \checkmark &             & 68.4(\textcolor{red}{+9.1})  & 81.7 & 85.7 & 58.5(\textcolor{red}{+8.5})  \\
            7     & \checkmark & \checkmark     & \checkmark & \checkmark &            & \checkmark  & 67.8(\textcolor{red}{+8.5})  & 80.9 & 85.0 & 57.8(\textcolor{red}{+7.8})  \\
            8     & \checkmark & \checkmark     & \checkmark & \checkmark & \checkmark &             & 70.3(\textcolor{red}{+11.0}) & 82.6 & 86.8 & 60.2(\textcolor{red}{+10.2}) \\
            9     & \checkmark & \checkmark     & \checkmark & \checkmark & \checkmark & \checkmark  & 71.3(\textcolor{red}{+12.0}) & 84.1 & 87.1 & 60.9(\textcolor{red}{+10.9}) \\
            \hline
            \hline
        \end{tabular}
        \label{ablation study}
    \end{table*}

    \subsection{Ablation Study}
    \label{ablation_section}

    In this part, we perform ablation studies on Occluded-Duke dataset to futher analyze the effectiveness of each component.

    \textbf{Effectiveness of proposed Modules.} we present the ablation studies of Dual-structure, Occlusion Instance Augmentation (OIA) and Feature Completion Decoder(FCD).  Table \ref{ablation study} shows the experimental results. Index-1 denotes the baseline transformer structure \cite{Transreid}. (1) From index-1 and index-2, we can see that the OIA module can improve 4.1\% Rank-1 accuracy and 3.2\% mAP compared with baseline model. OIA successfully incorporates a wider occlusion ratio and a variety of occlusion samples into the training set. (2) From index-2 and index-3, when dual-structure is employed, the dual-structure improves performance by 3.1\% Rank-1 accuracy and 2.5\% mAP. This demonstrates that dual-stream structure can effectively enable the shared part to learn a general occlusion and non-occlusion pattern, while the non-shared part completes the training of specific tasks. (3) The performance gain brought by FCD can be demonstrated by index-6 and index-8. After adding FCD to index-6, the performance increased by 1.9\% and 1.7\% on rank-1 and mAP respectively. By comparing index-3 and index-5, FCD could increase the performance by 0.8\% Rank-1 and 1.2\% mAP. This shows that the features obtained by the holistic branch are more robust with the help of CHT loss, making the supervision data of FCD more precise and the completion features produced by FCD more discriminative.

    \textbf{Effectiveness of proposed Losses.} we also present the ablation studies of Cross Hard Triplet Loss (CHT loss) and Feature Completion Consistency Loss (FC$^2$ loss). (1) Index-3 and index-6 show that the proposed CHT loss can contribute to network, improving the performance by 1.9\% Rank-1 and 2.8\% mAP. The CHT loss allows it to find the most difficult features that have different modalities than the anchors in the fused features. This is advantageous for attracting features with the same identity but distinct modalities in the feature space. (2) Both (Index-5, Index-7) and (Index-8, index-9) demonstrate that FC$^2$ loss can guide FCD to bring the generated pedestrian feature distribution closer to the holistic feature distribution, enabling FCD to complete pedestrian features while inputting occlusion features.

    \begin{figure}
        \centering
        \includegraphics[width=0.5\textwidth]{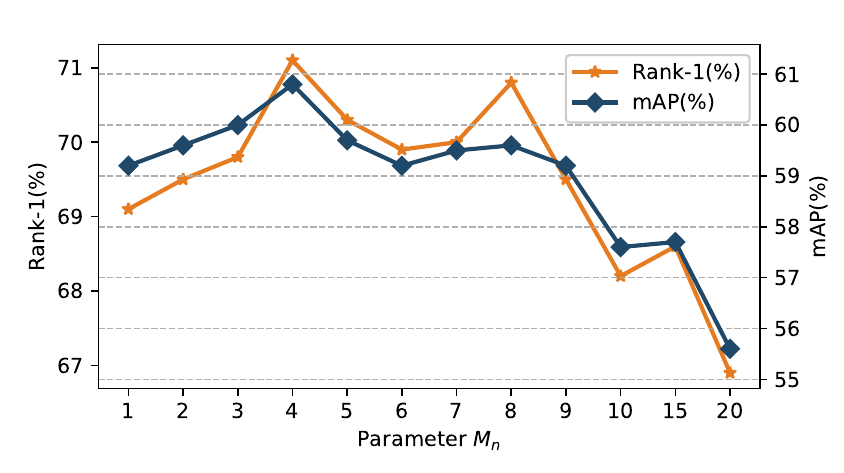}\\  
        \captionsetup{font={footnotesize}}
        \caption{Parameter analysis for the number of local features.}\label{Parts_ratio}
    \end{figure}

    \begin{figure}
        \centering
        \includegraphics[width=0.5\textwidth]{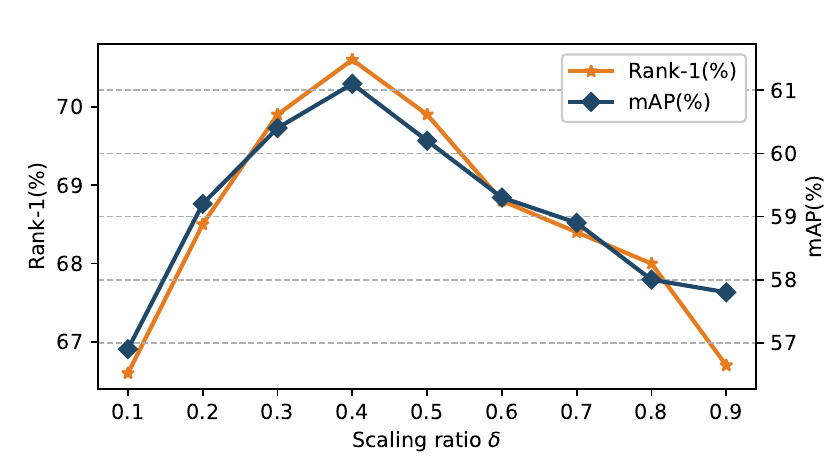}\\   
        \captionsetup{font={footnotesize}}
        \caption{Parameter analysis for the scaling ratio $\delta$ in OIA.}\label{OIA_ratio}
    \end{figure}

    \textbf{Analysis of the number of local features $M_n$.} The number of local feature determines the granularity of final output features. Fig. \ref{Parts_ratio} shows the performance of the REID influenced by the number of local features $M_n$. As we can see, FCFormer achieves the best Rank-1/mAP when $M_n = 4$. The initial performance improves as $M_n$ increases, which shows adequate local features can effectively help the model learn more robust occlusion features. However, increasing its value further contributes to performance decline. We conclude from Fig. \ref{Parts_ratio} that redundant tokens increase the dimensionality of features, and the model pays attention to more details, which weakens the anti-noise ability.

\begin{figure}[htbp]
        \centering
        \subfloat[]{
            \includegraphics[width=0.5in, height=0.9in]{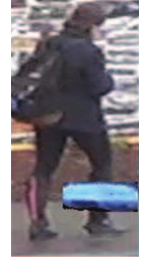}
            \label{small}
        }
        \centering
        \subfloat[]{
            \includegraphics[width=0.5in, height=0.9in]{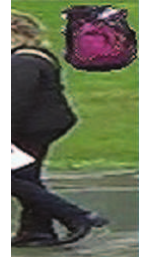}
            \label{failure}
        }%
        \centering
        \subfloat[]{
            \includegraphics[width=0.5in, height=0.9in]{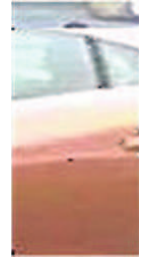}
            \label{large}
        }
        \centering
        \captionsetup{font={footnotesize}}
        \caption{The augmented image of (a) small scaling ratio. (b) failure case. (c) large scaling ratio.}
        \label{fail_exam}
    \end{figure}

    \textbf{Analysis of scaling ratio $\delta$ in OIA.} Fig. \ref{OIA_ratio} depicts the experimental findings of FCFormer at various fixed occlusion ratios. We note that the scaling ratio of 0.4 allows the FCFormer to perform at its best. Fig. \ref{fail_exam} shows that OIA may generate some undesired augmented images. For example, large portions of the human body's information are preserved in the augmented images produced by tiny occlusion ratios (Fig. \ref{small}). As a result, the network is unable to learn the relationship between person and occluder effectively. Another scenario about the small scaling ratio is that the occlusion sample does not hit the target person, resulting in an invalid augmentation (Fig. \ref{failure}). On the contrary, a high occlusion ratio ($\delta > 0.8$) (Fig. \ref{large}) results in a significant loss of pedestrian information, preventing the model from learning ID-related properties. Therefore, to imitate the real occlusion situation and prevent people from being entirely obscured, we take a uniform distribution centered around $\delta = 0.4$ ($\delta \sim \mathcal{U}(0.1, 0.7)$) instead of a fixed scaling ratio to scale the size of occlusion samples in \ref{SectionA}.

    \textbf{Analysis of augmentation locations in OIA.} Table \ref{OIA_position} shows that OIA uses different position enhancement methods. The first involves pasting images for each instance in a batch at a ``Fixed" location, and the second involves pasting images for each instance in a batch at ``Random" locations. As mentioned in \ref{SectionC}, our FCD adapts MAE's notion \cite{he2022masked} of restoring entire features using implicit unoccluded features. According to the results, the "random" enhancement approach can perform better than the "fixed" method. This is because the "fixed" method only selects a random position at the beginning, and then every instance in the batch follows this position, which reduces the diversity of occlusion relationships.

    \begin{table}[t]
        \small
        \centering
        \caption{Different augmentation methods within a batch.}
        \begin{tabular}{c|cccc}
            \hline
            Augmentation & R-1  & R-5  & R-10 & mAP  \\
            \hline
            \hline
            Fixed        & 67.8 & 83.2 & 86.1 & 57.4 \\
            Random       & 71.3 & 84.1 & 87.1 & 60.9 \\
            \hline
        \end{tabular}
        \label{OIA_position}
    \end{table}

    \begin{table}[htbp]
        \small
        \centering
        \caption{Comparison of different backbones. $*$ denotes that camera perspective information is introduced. $\S$ indicates that only the encoder is used to extract features.}
        \begin{tabular}{cc|cccc}
            \hline
            Backbone   & Param. & R-1           & R-5           & R-10          & mAP           \\
            \hline
            \hline
            ResNet50$^\S$ & -      & 42.6          & 57.1          & 62.9          & 33.7          \\
            ResNet50   & 25.6M  & 47.6          & 64.3          & 71.1          & 35.6          \\
            ResNet101  & 44.7M  & 47.7          & 62.5          & 68.7          & 35.6          \\
            \hline
            ViT-B$^\S$   & -      & 59.3          & 76.5          & 82.2          & 50.0          \\
            ViT-S*     & 22M    & 63.7          & 77.9          & 82.9          & 53.3          \\
            ViT-B*     & 86M    & \textbf{71.3} & \textbf{84.1} & \textbf{87.1} & \textbf{60.9} \\
            ViT-L*     & 307M   & 68.8          & 79.9          & 83.9          & 57.9          \\
            DeiT-B*    & 86M    & 69.5          & 82.9          & 86.1          & 59.7          \\
            \hline
            \hline
        \end{tabular}
        \label{different_backbone}
    \end{table}

    \begin{table}[t]
        \small
        \centering
        \caption{Influence on different augmentation methods.}
        \begin{tabular}{c|cccc}
            \hline
            Augmentation & R-1  & R-5  & R-10 & mAP  \\
            \hline
            \hline
            Random Erase        & 67.4 & 80.6 & 83.2 & 56.9 \\
            Mixup       & 66.7 & 79.6 & 81.1 & 55.3 \\
            CutOut      & 67.8 & 80.2 & 83.9 & 57.1 \\
            CutMix      & 69.0 & 82.4 & 85.2 & 58.1 \\
            OIA(ours)   & 71.3 & 84.1 & 87.1 & 60.9 \\
            \hline
        \end{tabular}
        \label{AUG}
    \end{table}

    \textbf{Analysis of different backbones.} In this section, we conduct experiments to show how the encoder Influences the performance. CNN-based and Transformer-based backbones are compared in Table \ref{different_backbone}. $\S$ means only the backbone is used for training. From the table, we can observer that the ResNet series and the Transformer series have a huge performance gap in occlusion scenarios. Here ResNet-50$^\S$ is similar to the PCB \cite{sun2018beyond} that directly learns local features. The second line shows that the FCFormer method takes ResNet-50 as backbone, surpassing ResNet$^\S$ by 5\% Rank-1 accuracy and 1.9\% mAP. It demonstrates that our proposed paradigm is effective. Unlike ResNet, Transformer has a high performance starting point when dealing with occlusion problems. ViT-Small has better performance and less parameters compared to ResNet-50. DeiT-B achieves the comparable results compared with ViT-B, the reason is that ViT series are pre-trained on ImageNet-21K and then finetuned on ImageNet-1K, but DeiT series only pre-trained on ImageNet-1K. The performance of ViT-L is lower than ViT-B, which shows that training large models on small dataset can easily lead to overfitting.

    \textbf{Influence on different augmentations.} In table \ref{AUG}, we verify the impact of different enhancement methods on the model (including Random Erase, Mixup, CutOut, CutMix, and OIA). The augmentation methods of Random Erase and CutOut are similar, that is, randomly selecting an area from the image to erase, so the performance of the two methods is equivalent.

    It is worth noting that the essence of Mixup is the fusion of two images, which does not produce a substantial occlusion relationship, so an occluded-holistic image pair cannot be obtained. However, we still designed an experiment to generate a holistic-mixup image pair, forcing the model recovers the original image from the mixup image. Relevant experiments show that FCFormer can achieve good performance with the enhancement of Mixup, but because it cannot learn the occlusion relationship, the model cannot obtain the best performance.

    CutMix augmentation has certain similarities with our augmentation, that is, cutting out pixels in one area and randomly filling pixels from other areas. Therefore, this augmentation can obtain competitive results, even surpassing some SOTA methods, indicating that our training paradigm is general and effective.

    \begin{table}[h]
        \small
        \centering
        \caption{Influence of occlusion learning paradigm.}
        \begin{tabular}{c|cccc}
            \hline
            Occlusion learning paradigm & R-1  & R-5  & R-10 & mAP  \\
            \hline
            \hline
            Implicitly        & 71.3 & 84.1 & 87.1 & 60.9 \\
            Explicitly        & 69.6 & 82.8 & 85.9 & 59.4 \\
            \hline
        \end{tabular}
        \label{occ_learning_paradigm}
    \end{table}

    \textbf{Occlusion learning paradigm.} To better explore the impact of implicitly learning occlusion or explicitly learning occlusion on the model, we design an experiment in which a binary classifier was added to the occlusion branch to determine whether it is occlusion. During testing, only when the occlusion classifier considers image to be occluded, the feature will be sent to the feature completion branch. The experimental results are shown in Table \ref{occ_learning_paradigm}. Since the shape and size of the occlusion changes greatly, the boundary defining the occlusion is relatively fuzzy, so the classifier is prone to overfitting. On the other hand, when there is no occlusion, the explicit method will not send the features to the feature completion branch. However, the implicit method can still bring positive benefits.

\subsection{Qualitative Analysis}
In this section, we present qualitative experimental results and demonstrate the superiority of our proposed FCFormer.

\begin{figure*}[t]
    \centering
    \subfloat[Baseline (ID loss)]{
        \includegraphics[width=0.31\textwidth]{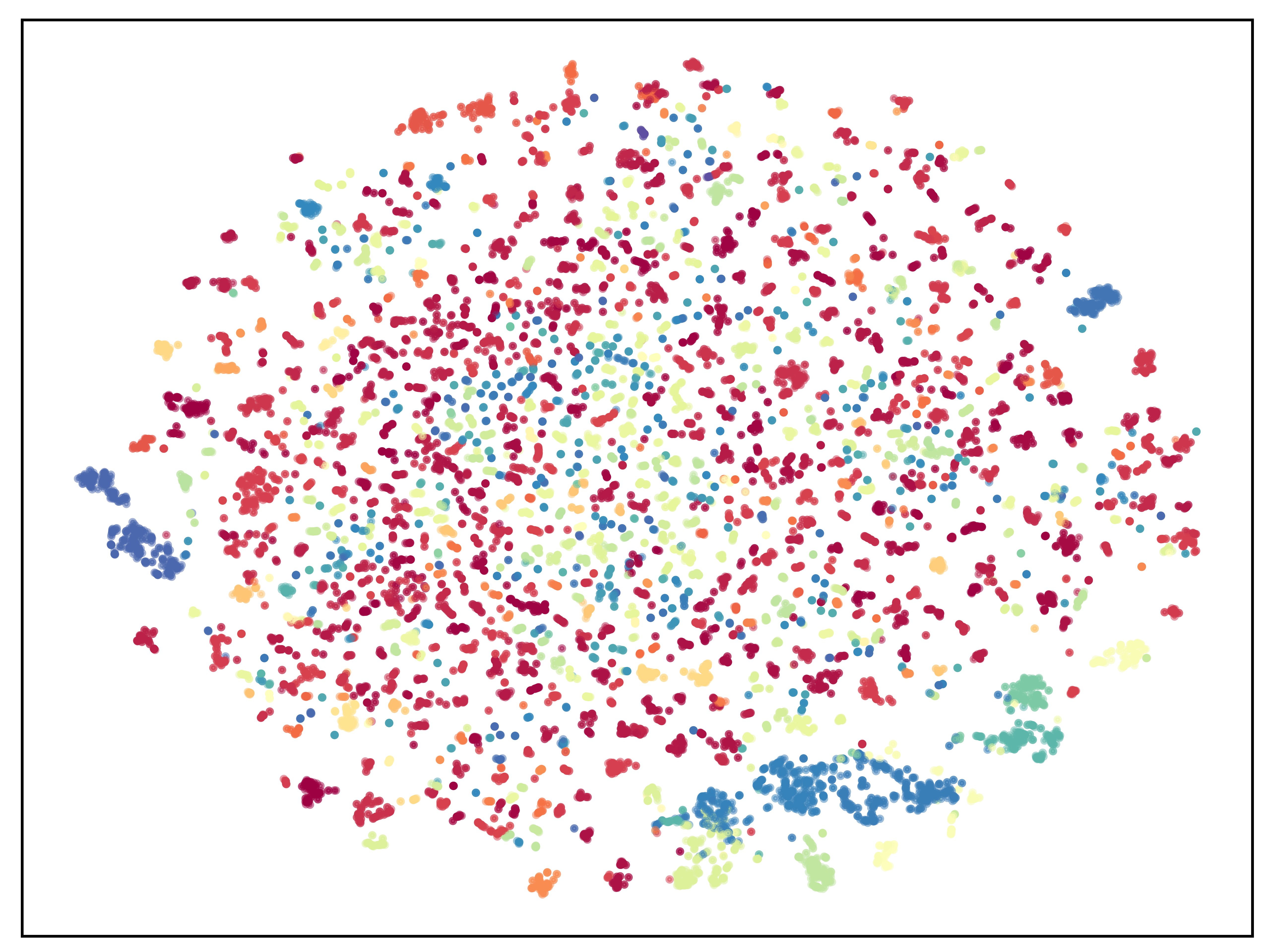}
        \label{tsne_a}
        }
    \subfloat[FCT (Triplet + ID loss)]{
        \includegraphics[width=0.31\textwidth]{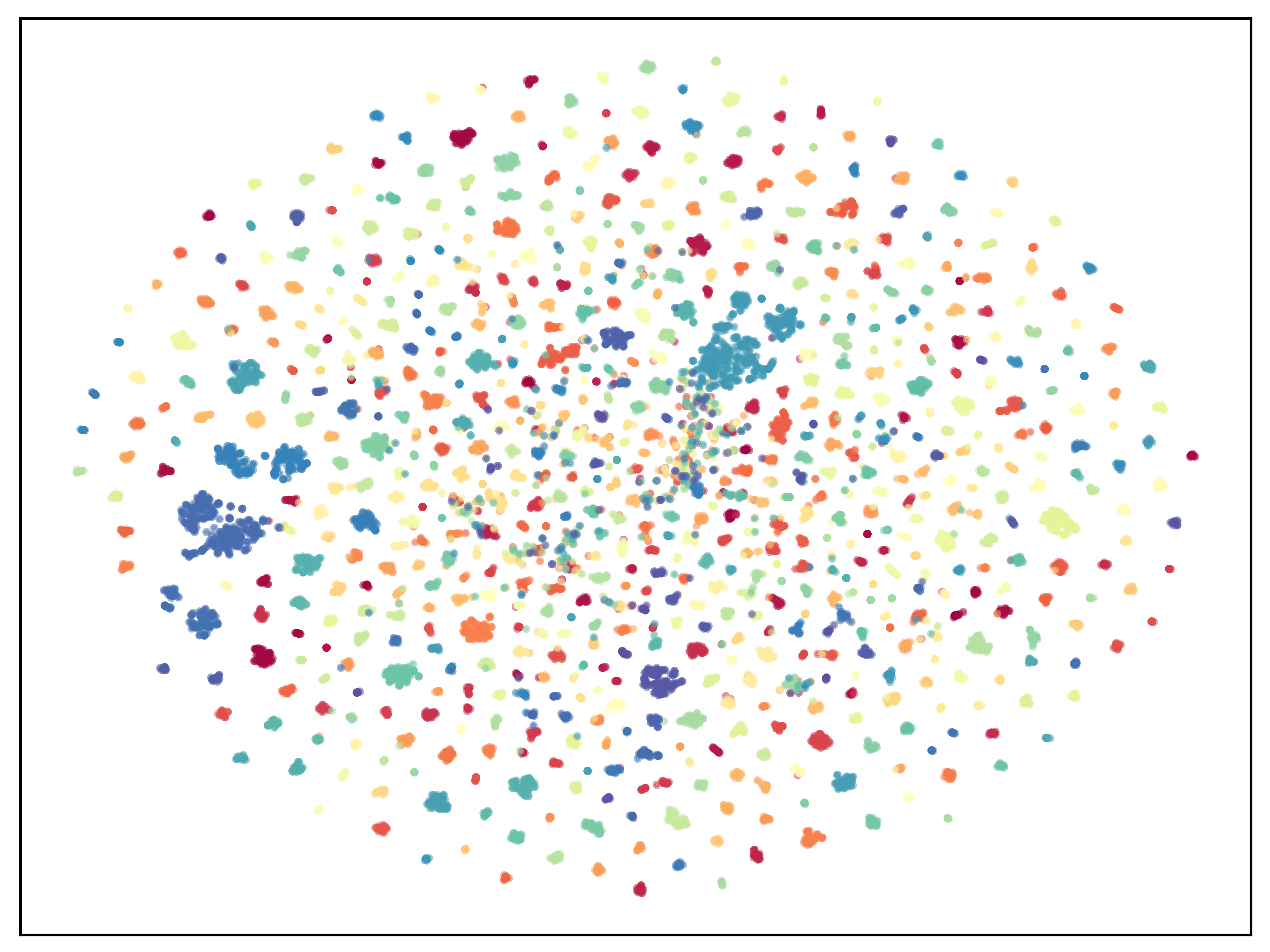}
        \label{tsne_b}
        }
    \subfloat[FCT (CHT + ID loss)]{
        \includegraphics[width=0.31\textwidth]{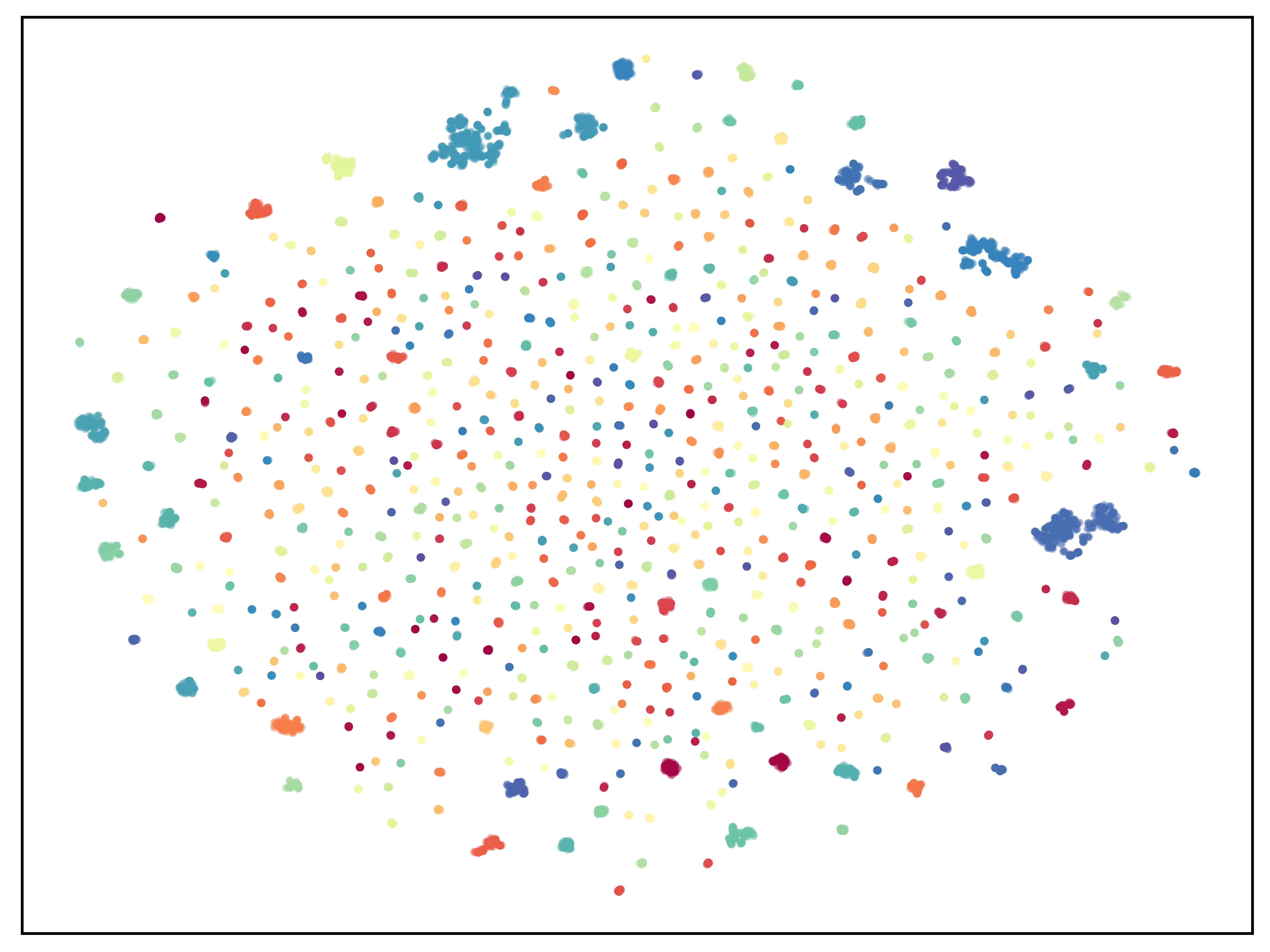}
        \label{tsne_c}
        }
    \captionsetup{font={footnotesize}}
    \caption{The t-SNE visualization of features. (a) denotes the feature distribution of baseline model trained only under ID loss. (b) denotes the feature distribution of FCT trained under Triplet + ID loss. (c) denotes the feature distribution of FCT trained under CHT + ID loss.}
    \label{tsne}
\end{figure*}

\textbf{Visualization of the feature distribution} We adopt t-SNE to show the distribution of the output features of the model under different constraints in Fig. \ref{tsne}. The output features are concated from occluded features, holistic features and completion features. As observed in this figure, the baseline model without triplet loss cannot cluster embeddings with the same ID well in occluded scenes. From Fig. \ref{tsne_b}, it can be seen that under the joint constraints of triplet and id loss, the model has been able to cluster better, but confusion still occurs under individual occlusion IDs. The results in Fig. \ref{tsne_c} show that the proposed CHT loss can better cluster the same ID more compactly in occlusion scenarios, which verifies that our method can learn invariant features well in complete, occluded and completed modes.


\begin{figure}[t]
    \centering
    \includegraphics[width=0.49\textwidth]{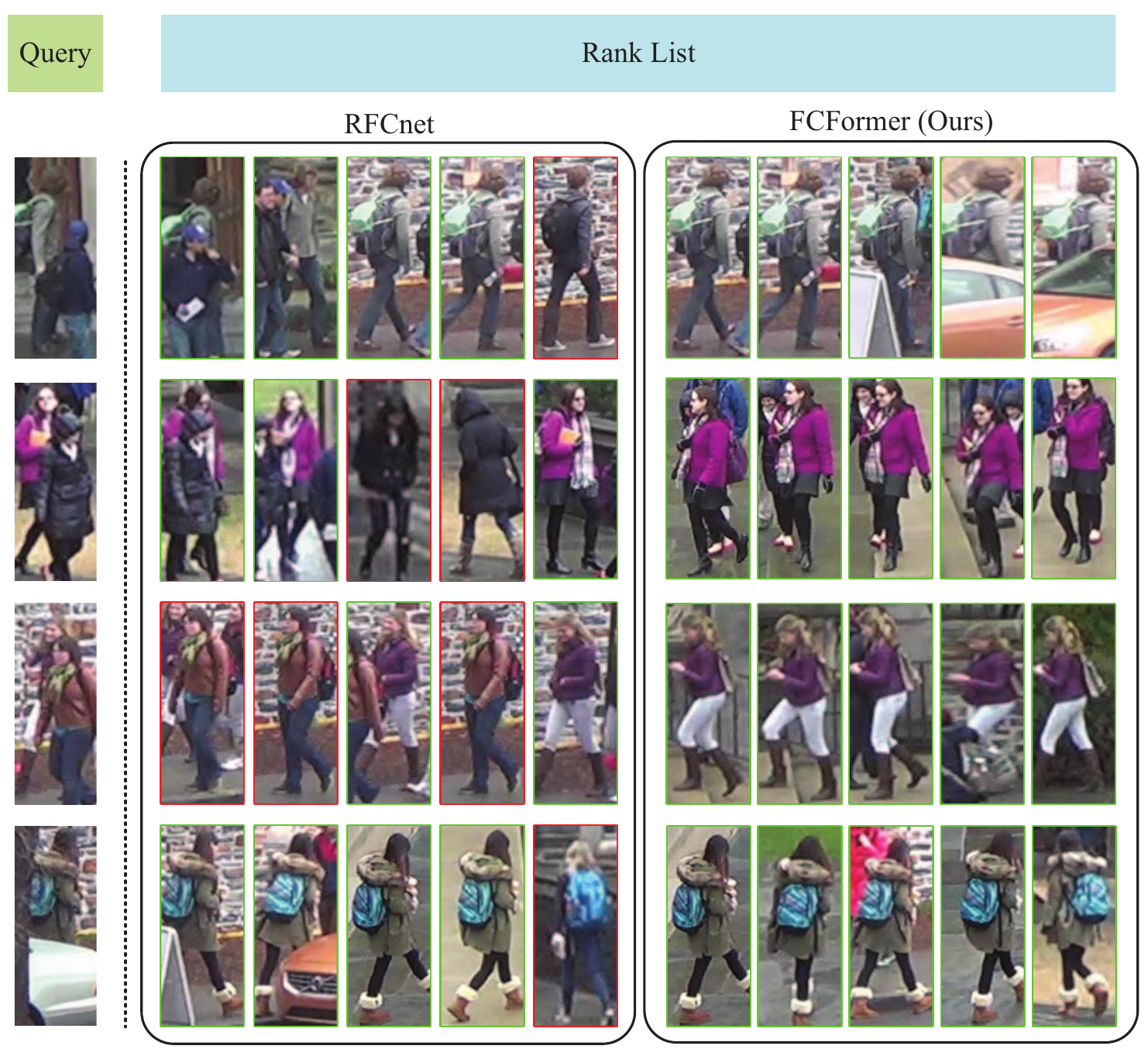}\\  
    \captionsetup{font={footnotesize}}
    \caption{Retrieval results of RFCnet and our proposed FCFormer on Occluded-DukeMTMC dataset.}
    \label{Rank_list}
\end{figure}

\textbf{Retrieval results of the feature completion transformer.} In Fig. \ref{Rank_list}, we present the retrieval results of RFCnet \cite{RFCnet} and our FCFormer.  The retrieval results show that RFCnet is prone to mix the information of the target person and obstacles, resulting in retrieving a wrong person with similar obstacle or treating the front person as target. This is because pre-designing the regions still does not tackle the extreme occlusion problem well. Unlike RFCnet's completion approach, we implicitly learn the completion features from the occlusion features without encoding and decoding on the feature space map.

\begin{figure}[h]
    \centering
    \includegraphics[width=0.49\textwidth]{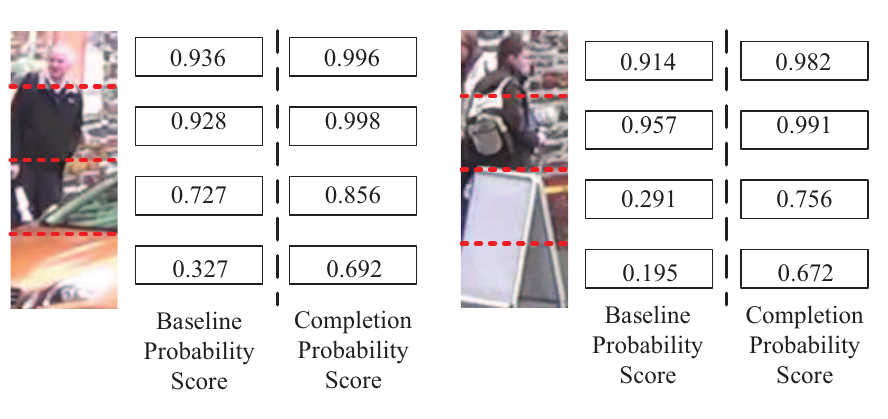}\\  
    \captionsetup{font={footnotesize}}
    \caption{Probability scores of local features}\label{Qualitative}
\end{figure}

\textbf{Probability scores of local features.} We present the probability scores after the non-shared transformer layers for some occluded pedestrian images. As shown in Fig. \ref{Qualitative}, occlusion heavily affects the probability scores of local features in the baseline model. However, the recovered local features obtained from FCD have significantly improved scores in occluded regions, which indicates that the feature completion decoder uses unoccluded information to compensate occluded region features.

\begin{figure}[h]
    \centering
    \includegraphics[width=0.48\textwidth]{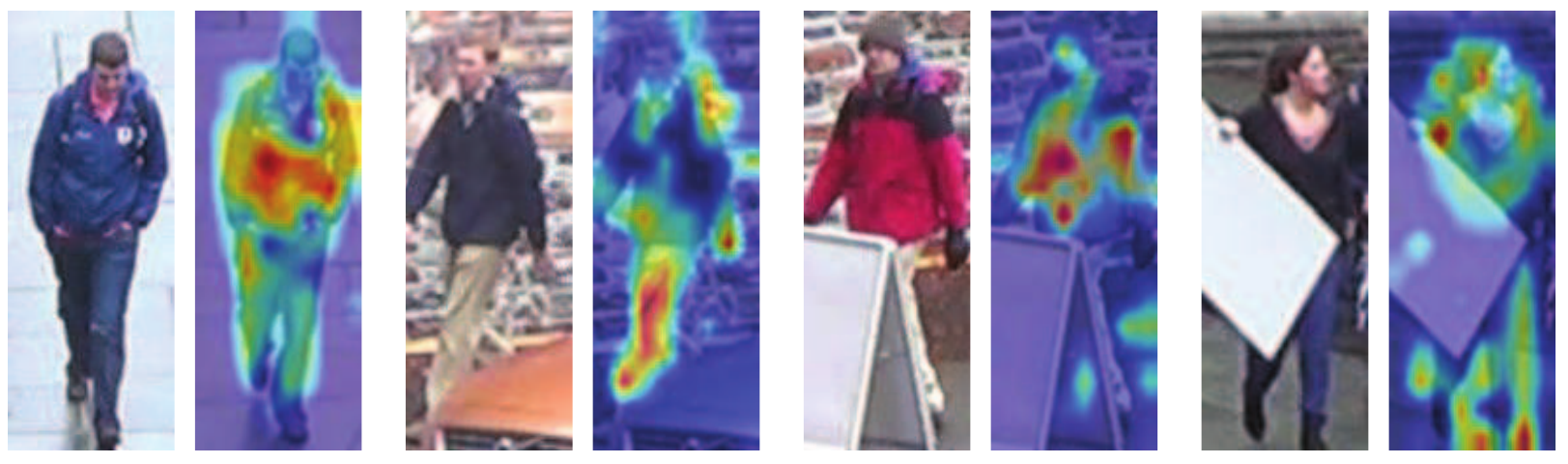}\\
    \captionsetup{font={footnotesize}}
    \caption{Visualization of attention heatmaps by Grad-CAM \cite{grad-cam} in feature completion decoder.}\label{GradCAM}
\end{figure}

\textbf{Visualization of attention maps in feature completion decoder.} We visualize the attention heatmaps for feature completion decoder in Fig. \ref{GradCAM} by using Grad-CAM \cite{grad-cam}. The model concentrates on body parts with significant information when the pedestrian is not occluded. On the contrary, if the target person is occluded, FCD will pay attention to the position of the human body on the neighbor region of occluder, as this helps to complement missing features with nearby features.

\section{Conclusion}
In this paper, we propose a Feature Completion Transformer (FCFormer) framework to alleviate occluded person re-identification problem. The core idea is to construct a variety of occlusion sample pairs, and train a decoder that can complement incomplete pedestrian features through known occlusion relationships without relying on extra semantic model. Specifically, OIA provides a new occlusion augmentation scheme with generality, providing a variety of holistic-occlusion sample pairs. Subsequently, a dual stream structure can use the shared encoder to train two different branches for holistic and occluded images respectively by using holistic-occlusion sample pairs. Then we propose a feature completion decoder to recovery holistic features from occluded features. Extensive experiments on occluded and holistic datasets demonstrate the effectiveness of our proposed method.




\bibliographystyle{IEEEtran}

\begin{thebibliography}{10}
    \providecommand{\url}[1]{#1}
    \csname url@samestyle\endcsname
    \providecommand{\newblock}{\relax}
    \providecommand{\bibinfo}[2]{#2}
    \providecommand{\BIBentrySTDinterwordspacing}{\spaceskip=0pt\relax}
    \providecommand{\BIBentryALTinterwordstretchfactor}{4}
    \providecommand{\BIBentryALTinterwordspacing}{\spaceskip=\fontdimen2\font plus
    \BIBentryALTinterwordstretchfactor\fontdimen3\font minus
      \fontdimen4\font\relax}
    \providecommand{\BIBforeignlanguage}[2]{{%
    \expandafter\ifx\csname l@#1\endcsname\relax
    \typeout{** WARNING: IEEEtran.bst: No hyphenation pattern has been}%
    \typeout{** loaded for the language `#1'. Using the pattern for}%
    \typeout{** the default language instead.}%
    \else
    \language=\csname l@#1\endcsname
    \fi
    #2}}
    \providecommand{\BIBdecl}{\relax}
    \BIBdecl
    
    \bibitem{person_reid}
    L.~Zheng, Y.~Yang, and A.~G. Hauptmann, ``Person re-identification: Past,
      present and future,'' \emph{arXiv preprint arXiv:1610.02984}, 2016.
    
    \bibitem{sun2018beyond}
    Y.~Sun, L.~Zheng, Y.~Yang, Q.~Tian, and S.~Wang, ``Beyond part models: Person
      retrieval with refined part pooling (and a strong convolutional baseline),''
      in \emph{Proceedings of the European Conference on Computer Vision (ECCV)},
      2018, pp. 480--496.
    
    \bibitem{shi2022image}
    W.~Shi, H.~Liu, and M.~Liu, ``Image-to-video person re-identification using
      three-dimensional semantic appearance alignment and cross-modal interactive
      learning,'' in \emph{Pattern Recognition}.\hskip 1em plus 0.5em minus
      0.4em\relax Elsevier, 2022, p. 108314.
    
    \bibitem{shi2020identity}
    W.~Shi, H.~Liu, and M.~Liu, ``Identity-sensitive loss guided and instance feature boosted deep
      embedding for person search,'' in \emph{Neurocomputing}, vol. 415, 2020, pp.
      1--14.
    
    \bibitem{zhuo2018occluded}
    J.~Zhuo, Z.~Chen, J.~Lai, and G.~Wang, ``Occluded person re-identification,''
      in \emph{2018 IEEE International Conference on Multimedia and Expo (ICME)},
      2018, pp. 1--6.
    
    \bibitem{miao2019pose}
    J.~Miao, Y.~Wu, P.~Liu, Y.~Ding, and Y.~Yang, ``Pose-guided feature alignment
      for occluded person re-identification,'' in \emph{Proceedings of the IEEE/CVF
      International Conference on Computer Vision (ICCV)}, 2019, pp. 542--551.
    
    \bibitem{OAMN}
    P.~Chen, W.~Liu, P.~Dai, J.~Liu, Q.~Ye, M.~Xu, Q.~Chen, and R.~Ji, ``Occlude
      them all: Occlusion-aware attention network for occluded person re-id,'' in
      \emph{Proceedings of the IEEE/CVF international conference on computer vision
      (CVPR)}, 2021, pp. 11\,833--11\,842.
    
    \bibitem{DRL_Net}
    M.~Jia, X.~Cheng, S.~Lu, and J.~Zhang, ``Learning disentangled representation
      implicitly via transformer for occluded person re-identification,''
      \emph{IEEE Transactions on Multimedia (TMM)}, 2022.
    
    \bibitem{PVPM}
    S.~Gao, J.~Wang, H.~Lu, and Z.~Liu, ``Pose-guided visible part matching for
      occluded person reid,'' in \emph{Proceedings of the IEEE/CVF Conference on
      Computer Vision and Pattern Recognition (CVPR)}, 2020, pp. 11\,744--11\,752.
    
    \bibitem{HOReID}
    G.~Wang, S.~Yang, H.~Liu, Z.~Wang, Y.~Yang, S.~Wang, G.~Yu, E.~Zhou, and
      J.~Sun, ``High-order information matters: Learning relation and topology for
      occluded person re-identification,'' in \emph{Proceedings of the IEEE/CVF
      Conference on Computer Vision and Pattern Recognition (CVPR)}, 2020, pp.
      6449--6458.
    
    \bibitem{GAN_Part}
    S.~Iodice and K.~Mikolajczyk, ``Partial person re-identification with alignment
      and hallucination,'' in \emph{Computer Vision--ACCV 2018: 14th Asian
      Conference on Computer Vision (ACCV)}.\hskip 1em plus 0.5em minus 0.4em\relax
      Springer, 2019, pp. 101--116.
    
    \bibitem{GAN_semantic}
    X.~Jin, C.~Lan, W.~Zeng, G.~Wei, and Z.~Chen, ``Semantics-aligned
      representation learning for person re-identification,'' in \emph{Proceedings
      of the AAAI Conference on Artificial Intelligence (AAAI)}, vol.~34, no.~07,
      2020, pp. 11\,173--11\,180.
    
    \bibitem{RFCnet}
    R.~Hou, B.~Ma, H.~Chang, X.~Gu, S.~Shan, and X.~Chen, ``Feature completion for
      occluded person re-identification,'' \emph{IEEE Transactions on Pattern
      Analysis and Machine Intelligence (TPAMI)}, vol.~44, no.~9, pp. 4894--4912,
      2021.
    
    \bibitem{COCO}
    T.-Y. Lin, M.~Maire, S.~Belongie, J.~Hays, P.~Perona, D.~Ramanan,
      P.~Doll{\'a}r, and C.~L. Zitnick, ``Microsoft coco: Common objects in
      context,'' in \emph{European Conference on Computer Vision (ECCV)}, 2014, pp.
      740--755.
    
    \bibitem{Random}
    Z.~Zhong, L.~Zheng, G.~Kang, S.~Li, and Y.~Yang, ``Random erasing data
      augmentation,'' \emph{Proceedings of the AAAI Conference on Artificial
      Intelligence (AAAI)}, vol.~34, no.~7, 2017.
    
    \bibitem{alignedreid}
    X.~Zhang, H.~Luo, X.~Fan, W.~Xiang, Y.~Sun, Q.~Xiao, W.~Jiang, C.~Zhang, and
      J.~Sun, ``Alignedreid: Surpassing human-level performance in person
      re-identification,'' \emph{arXiv preprint arXiv:1711.08184}, 2017.
    
    \bibitem{vpm}
    Y.~Sun, Q.~Xu, Y.~Li, C.~Zhang, Y.~Li, S.~Wang, and J.~Sun, ``Perceive where to
      focus: Learning visibility-aware part-level features for partial person
      re-identification,'' in \emph{Proceedings of the IEEE/CVF Conference on
      Computer Vision and Pattern Recognition (CVPR)}, 2019, pp. 393--402.
    
    \bibitem{MoS}
    M.~Jia, X.~Cheng, Y.~Zhai, S.~Lu, S.~Ma, Y.~Tian, and J.~Zhang, ``Matching on
      sets: Conquer occluded person re-identification without alignment,'' in
      \emph{Proceedings of the AAAI Conference on Artificial Intelligence (AAAI)},
      vol.~35, 2021, pp. 1673--1681.
    
    \bibitem{song2018mask}
    C.~Song, Y.~Huang, W.~Ouyang, and L.~Wang, ``Mask-guided contrastive attention
      model for person re-identification,'' in \emph{Proceedings of the IEEE
      Conference on Computer Vision and Pattern Recognition (CVPR)}, 2018, pp.
      1179--1188.
    
    \bibitem{Neighbourhood}
    S.~Yu, D.~Chen, R.~Zhao, H.~Chen, and Y.~Qiao, ``Neighbourhood-guided feature
      reconstruction for occluded person re-identification,'' \emph{arXiv preprint
      arXiv:2105.07345}, 2021.
    
    \bibitem{parallel}
    H.~Huang, A.~Zheng, C.~Li, R.~He \emph{et~al.}, ``Parallel augmentation and
      dual enhancement for occluded person re-identification,'' \emph{arXiv
      preprint arXiv:2210.05438}, 2022.
    
    \bibitem{matterport_maskrcnn_2017}
    K.~He, G.~Gkioxari, P.~Doll{\'a}r, and R.~Girshick, ``Mask r-cnn,'' in
      \emph{Proceedings of the IEEE international conference on computer vision
      (ICCV)}, 2017, pp. 2961--2969.
    
    \bibitem{Transreid}
    S.~He, H.~Luo, P.~Wang, F.~Wang, H.~Li, and W.~Jiang, ``Transreid:
      Transformer-based object re-identification,'' in \emph{Proceedings of the
      IEEE/CVF International Conference on Computer Vision (ICCV)}, 2021.
    
    \bibitem{PFD}
    T.~Wang, H.~Liu, P.~Song, T.~Guo, and W.~Shi, ``Pose-guided feature
      disentangling for occluded person re-identification based on transformer,''
      in \emph{Proceedings of the AAAI Conference on Artificial Intelligence
      (AAAI)}, vol.~36, no.~3, 2022, pp. 2540--2549.
    
    \bibitem{dosovitskiy2020image}
    A.~Dosovitskiy, L.~Beyer, A.~Kolesnikov, D.~Weissenborn, X.~Zhai,
      T.~Unterthiner, M.~Dehghani, M.~Minderer, G.~Heigold, S.~Gelly \emph{et~al.},
      ``An image is worth 16x16 words: Transformers for image recognition at
      scale,'' \emph{arXiv preprint arXiv:2010.11929}, 2020.
    
    \bibitem{PAT}
    Y.~Li, J.~He, T.~Zhang, X.~Liu, Y.~Zhang, and F.~Wu, ``Diverse part discovery:
      Occluded person re-identification with part-aware transformer,'' in
      \emph{Proceedings of the IEEE/CVF Conference on Computer Vision and Pattern
      Recognition (CVPR)}, 2021, pp. 2898--2907.
    
    \bibitem{BN}
    S.~Ioffe and C.~Szegedy, ``Batch normalization: Accelerating deep network
      training by reducing internal covariate shift,'' in \emph{International
      conference on machine learning}.\hskip 1em plus 0.5em minus 0.4em\relax pmlr,
      2015, pp. 448--456.
    
    \bibitem{he2022masked}
    K.~He, X.~Chen, S.~Xie, Y.~Li, P.~Doll{\'a}r, and R.~Girshick, ``Masked
      autoencoders are scalable vision learners,'' in \emph{Proceedings of the IEEE
      conference on computer vision and pattern recognition (CVPR)}, 2022, pp.
      16\,000--16\,009.
    
    \bibitem{vaswani2017attention}
    A.~Vaswani, N.~Shazeer, N.~Parmar, J.~Uszkoreit, L.~Jones, A.~N. Gomez,
      {\L}.~Kaiser, and I.~Polosukhin, ``Attention is all you need,'' in
      \emph{Advances in Neural Information Processing Systems (NeurIPS)}, 2017, pp.
      5998--6008.
    
    \bibitem{schroff2015facenet}
    F.~Schroff, D.~Kalenichenko, and J.~Philbin, ``Facenet: A unified embedding for
      face recognition and clustering,'' in \emph{Proceedings of the IEEE
      conference on computer vision and pattern recognition (CVPR)}, 2015, pp.
      815--823.
    
    \bibitem{FED}
    Z.~Wang, F.~Zhu, S.~Tang, R.~Zhao, L.~He, and J.~Song, ``Feature erasing and
      diffusion network for occluded person re-identification,'' in
      \emph{Proceedings of the IEEE/CVF Conference on Computer Vision and Pattern
      Recognition (CVPR))}, 2022, pp. 4754--4763.
    
    \bibitem{zhao2017deeply}
    L.~Zhao, X.~Li, Y.~Zhuang, and J.~Wang, ``Deeply-learned part-aligned
      representations for person re-identification,'' in \emph{Proceedings of the
      IEEE/CVF International Conference on Computer Vision (ICCV)}, 2017, pp.
      3219--3228.
    
    \bibitem{PartBilinear}
    Y.~Suh, J.~Wang, S.~Tang, T.~Mei, and K.~M. Lee, ``Part-aligned bilinear
      representations for person re-identification,'' in \emph{Proceedings of the
      European Conference on Computer Vision (ECCV)}, 2018, pp. 402--419.
    
    \bibitem{fd-gan}
    Y.~Ge, Z.~Li, H.~Zhao, G.~Yin, S.~Yi, X.~Wang, and H.~Li, ``Fd-gan: Pose-guided
      feature distilling gan for robust person re-identification,'' \emph{arXiv
      preprint arXiv:1810.02936}, 2018.
    
    \bibitem{DSR}
    L.~He, J.~Liang, H.~Li, and Z.~Sun, ``Deep spatial feature reconstruction for
      partial person re-identification: Alignment-free approach,'' in
      \emph{Proceedings of the IEEE Conference on Computer Vision and Pattern
      Recognition (CVPR)}, 2018, pp. 7073--7082.
    
    \bibitem{SFR}
    L.~He, Z.~Sun, Y.~Zhu, and Y.~Wang, ``Recognizing partial biometric patterns,''
      \emph{arXiv preprint arXiv:1810.07399}, 2018.
    
    \bibitem{Ad-occ}
    H.~Huang, D.~Li, Z.~Zhang, X.~Chen, and K.~Huang, ``Adversarially occluded
      samples for person re-identification,'' in \emph{Proceedings of the IEEE
      Conference on Computer Vision and Pattern Recognition (CVPR)}, 2018, pp.
      5098--5107.
    
    \bibitem{ISP}
    K.~Zhu, H.~Guo, Z.~Liu, M.~Tang, and J.~Wang, ``Identity-guided human semantic
      parsing for person re-identification,'' in \emph{European Conference on
      Computer Vision (ECCV)}, 2020, pp. 346--363.
    
    \bibitem{SORN}
    X.~Zhang, Y.~Yan, J.-H. Xue, Y.~Hua, and H.~Wang, ``Semantic-aware
      occlusion-robust network for occluded person re-identification,'' \emph{IEEE
      Transactions on Circuits and Systems for Video Technology (TCSVT)}, vol.~31,
      no.~7, pp. 2764--2778, 2021.
    
    \bibitem{Pirt}
    Z.~Ma, Y.~Zhao, and J.~Li, ``Pose-guided inter-and intra-part relational
      transformer for occluded person re-identification,'' in \emph{Proceedings of
      the 29th ACM International Conference on Multimedia (ACM MM)}, 2021, pp.
      1487--1496.
    
    \bibitem{PGFL_KD2021}
    K.~Zheng, C.~Lan, W.~Zeng, J.~Liu, Z.~Zhang, and Z.-J. Zha, ``Pose-guided
      feature learning with knowledge distillation for occluded person
      re-identification,'' in \emph{Proceedings of the 29th ACM International
      Conference on Multimedia (ACM MM)}, 2021, pp. 4537--4545.
    
    \bibitem{QPM}
    P.~Wang, C.~Ding, Z.~Shao, Z.~Hong, S.~Zhang, and D.~Tao, ``Quality-aware part
      models for occluded person re-identification,'' \emph{IEEE Transactions on
      Multimedia (TMM)}, pp. 1--1, 2022.
    
    \bibitem{xu2022learning}
    B.~Xu, L.~He, J.~Liang, and Z.~Sun, ``Learning feature recovery transformer for
      occluded person re-identification,'' \emph{IEEE Transactions on Image
      Processing}, vol.~31, pp. 4651--4662, 2022.
    
    \bibitem{zheng2017unlabeled}
    Z.~Zheng, L.~Zheng, and Y.~Yang, ``Unlabeled samples generated by gan improve
      the person re-identification baseline in vitro,'' in \emph{Proceedings of the
      IEEE/CVF International Conference on Computer Vision (ICCV)}, 2017, pp.
      3754--3762.
    
    \bibitem{zheng2015scalable}
    L.~Zheng, L.~Shen, L.~Tian, S.~Wang, J.~Wang, and Q.~Tian, ``Scalable person
      re-identification: A benchmark,'' in \emph{Proceedings of the IEEE/CVF
      International Conference on Computer Vision (ICCV)}, 2015, pp. 1116--1124.
    
    \bibitem{spreid}
    M.~M. Kalayeh, E.~Basaran, M.~G{\"o}kmen, M.~E. Kamasak, and M.~Shah, ``Human
      semantic parsing for person re-identification,'' in \emph{Proceedings of the
      IEEE Conference on Computer Vision and Pattern Recognition (CVPR)}, 2018, pp.
      1062--1071.
    
    \bibitem{BOT}
    H.~Luo, Y.~Gu, X.~Liao, S.~Lai, and W.~Jiang, ``Bag of tricks and a strong
      baseline for deep person re-identification,'' in \emph{Proceedings of the
      IEEE/CVF Conference on Computer Vision and Pattern Recognition Workshops
      (CVPRW)}, 2019.
    
    \bibitem{MVPM}
    H.~Sun, Z.~Chen, S.~Yan, and L.~Xu, ``Mvp matching: A maximum-value perfect
      matching for mining hard samples, with application to person
      re-identification,'' in \emph{Proceedings of the IEEE/CVF International
      Conference on Computer Vision (ICCV)}, 2019, pp. 6737--6747.
    
    \bibitem{sft}
    C.~Luo, Y.~Chen, N.~Wang, and Z.~Zhang, ``Spectral feature transformation for
      person re-identification,'' in \emph{Proceedings of the IEEE/CVF
      International Conference on Computer Vision (ICCV)}, 2019, pp. 4976--4985.
    
    \bibitem{CAMA}
    W.~Yang, H.~Huang, Z.~Zhang, X.~Chen, K.~Huang, and S.~Zhang, ``Towards rich
      feature discovery with class activation maps augmentation for person
      re-identification,'' in \emph{Proceedings of the IEEE/CVF Conference on
      Computer Vision and Pattern Recognition (CVPR)}, 2019, pp. 1389--1398.
    
    \bibitem{IAnet}
    R.~Hou, B.~Ma, H.~Chang, X.~Gu, S.~Shan, and X.~Chen,
      ``Interaction-and-aggregation network for person re-identification,'' in
      \emph{Proceedings of the IEEE/CVF Conference on Computer Vision and Pattern
      Recognition (CVPR)}, 2019, pp. 9317--9326.
    
    \bibitem{p2net}
    J.~Guo, Y.~Yuan, L.~Huang, C.~Zhang, J.-G. Yao, and K.~Han, ``Beyond human
      parts: Dual part-aligned representations for person re-identification,'' in
      \emph{Proceedings of the IEEE/CVF International Conference on Computer Vision
      (ICCV)}, 2019, pp. 3642--3651.
    
    \bibitem{AAnet}
    C.-P. Tay, S.~Roy, and K.-H. Yap, ``Aanet: Attribute attention network for
      person re-identifications,'' in \emph{Proceedings of the IEEE/CVF Conference
      on Computer Vision and Pattern Recognition (CVPR)}, 2019, pp. 7134--7143.
    
    \bibitem{circle}
    Y.~Sun, C.~Cheng, Y.~Zhang, C.~Zhang, L.~Zheng, Z.~Wang, and Y.~Wei, ``Circle
      loss: A unified perspective of pair similarity optimization,'' in
      \emph{Proceedings of the IEEE/CVF Conference on Computer Vision and Pattern
      Recognition (CVPR)}, 2020, pp. 6398--6407.
    
    \bibitem{HACNN}
    W.~Li, X.~Zhu, and S.~Gong, ``Harmonious attention network for person
      re-identification,'' in \emph{Proceedings of the IEEE conference on computer
      vision and pattern recognition (CVPR)}, 2018, pp. 2285--2294.
    
    \bibitem{OSNet}
    K.~Zhou, Y.~Yang, A.~Cavallaro, and T.~Xiang, ``Omni-scale feature learning for
      person re-identification,'' in \emph{Proceedings of the IEEE/CVF
      International Conference on Computer Vision (ICCV)}, 2019, pp. 3702--3712.
    
    \bibitem{IDE}
    L.~Zheng, H.~Zhang, S.~Sun, M.~Chandraker, Y.~Yang, and Q.~Tian, ``Person
      re-identification in the wild,'' in \emph{Proceedings of the IEEE conference
      on computer vision and pattern recognition (CVPR)}, 2017, pp. 1367--1376.
    
    \bibitem{Part_bili}
    Y.~Suh, J.~Wang, S.~Tang, T.~Mei, and K.~M. Lee, ``Part-aligned bilinear
      representations for person re-identification,'' in \emph{Proceedings of the
      European conference on computer vision (ECCV)}, 2018, pp. 402--419.
    
    \bibitem{grad-cam}
    R.~R. Selvaraju, M.~Cogswell, A.~Das, R.~Vedantam, D.~Parikh, and D.~Batra,
      ``Grad-cam: Visual explanations from deep networks via gradient-based
      localization,'' in \emph{Proceedings of the IEEE international conference on
      computer vision (ICCV)}, 2017, pp. 618--626.
    
    \end{thebibliography}

\end{document}